\documentclass{bmvc2k}

\usepackage{amsmath}
\usepackage{amssymb}

\usepackage{graphicx}
\usepackage{microtype}
\usepackage{booktabs}       %
\usepackage{multirow}
\usepackage{wrapfig}
\usepackage{graphbox}
\usepackage{xcolor}
\usepackage{color}
\usepackage{microtype} 
\usepackage{booktabs} 
\usepackage{widetext}
\usepackage{amsthm}

\title{Robust Semantic Segmentation with Superpixel-Mix}

\addauthor{Gianni Franchi}{gianni.franchi@ensta-paris.fr}{1}
\addauthor{Nacim Belkhir}{nacim.belkhir@safrangroup.com}{2}
\addauthor{Mai Lan Ha}{hamailan@informatik.uni-siegen.de}{3}

\addauthor{Yufei Hu }{yufei.hu.2021@ensta-paris.fr}{1}
\addauthor{Andrei Bursuc}{andrei.bursuc@valeo.com}{4}
\addauthor{Volker Blanz}{blanz@informatik.uni-siegen.de}{3}
\addauthor{Angela Yao}{ayao@comp.nus.edu.sg}{5}

\newcommand{\ab}[1]{\textcolor{black}{#1}}
\newcommand\Correction{\textcolor{black}}

\newcommand{\AB}[1]{\textcolor{black}{#1}}

\addinstitution{
 U2IS ENSTA Paris\\
Institut Polytechnique de Paris
}
\addinstitution{
Safrantech, Safran Group
}
\addinstitution{
Department of Computer\\
 Science University of Siegen
}
\addinstitution{
valeo.ai
}

\addinstitution{
School of Computing\\
 National University of Singapore
}
\runninghead{Franchi \etal}{Robust Semantic Segmentation with Superpixel-Mix}

\def\etal{\emph{et al}\bmvaOneDot}

 \newcommand*{\ie}{\emph{i.e.}\@\xspace}

\begin{document}

\maketitle

\begin{abstract}

Along with predictive performance and runtime speed, \Correction{robustness} is a key requirement for real-world semantic segmentation. %
\Correction{Robustness encompasses accuracy, predictive uncertainty, stability under data  perturbation and distribution shift, and reduced bias.} To improve \Correction{robustness}, we introduce Superpixel-mix, a new superpixel-based data augmentation method with teacher-student consistency training. Unlike other mixing-based augmentation techniques, mixing superpixels between images 
is aware of object boundaries, while yielding consistent gains in segmentation accuracy. Our 
proposed technique achieves state-of-the-art results in semi-supervised semantic segmentation on the Cityscapes dataset. Moreover, Superpixel-mix 
improves the \Correction{robustness} of semantic segmentation by reducing network uncertainty and bias, as confirmed by competitive results under strong distributions shift (adverse weather, image corruptions) and when facing out-of-distribution data.
\end{abstract}

\section{Introduction}

Semantic segmentation is an important task in computer vision with a high potential for practical applications, in particular for autonomous vehicles. Thanks to the predicted 2D segmentation maps by the deep convolutional neural networks (DCNNs), it contributes to an improved understanding of the scenes.

A large body of the recent literature on DCNNs for semantic segmentation focuses on improving predictive performance and run-time through advanced~\cite{chen2017deeplab,yu2017dilated} or lighter~\cite{yu2018bisenet, zhao2018icnet, li2019dfanet, li2019partial} architectures, better use of multiple resolutions~\cite{zhao2017pyramid, sun2019high} and novel loss functions~\cite{lin2017focal, sudre2017generalised, berman2018lovasz, zhu2019improving}. However, for real-world deployments, other requirements, e.g., reliability, robustness, must be equally satisfied to avoid any failures. To reach them, there are a few major challenges yet to be fully solved. First, DCNNs have been shown to be overconfident~\cite{guo2017calibration} even when their predictions are wrong~\cite{nguyen2015deep, hein2019relu}. In addition, DCNNs struggle to learn when there are few training samples are available, or data and annotations are noisy. In particular, high capacity DCNNs can find ``shortcuts'' that allow them to exploit spurious correlations in the data (e.g., background information~\cite{rosenfeld2018elephant, srivasta2020human}, textures and salient patterns~\cite{geirhos2018imagenettrained}) towards minimizing the training error at the cost of generalization. Such DCNNs have been shown to be biased, e.g. contextual bias~\cite{xiao2021noise} or texture bias~\cite{geirhos2018imagenettrained}. This type of problem could be addressed by larger and higher quality datasets~\cite{lambert2020mseg}, yet the entire complexity of the world cannot be encompassed in a training dataset with a limited size. Alternative solutions leverage uncertainty estimation for detecting such failures~\cite{gal2016dropout, lakshminarayanan2017simple, franchi2019tradi}. However the most effective ones are computationally inefficient as they rely on ensembles or multiple forward passes~\cite{ovadia2019can, gustafsson2020evaluating, ashukha2020pitfalls}.

In this paper, we aim to increase the \Correction{robustness} of semantic segmentation models. For the scope of this work, we define \Correction{\emph{robust} as follows}: \emph{a model is \Correction{robust} if its predictions are accurate and well calibrated when facing 
typical conditions from the training distribution, but also under distribution shift (epistemic and aleatoric uncertainty) and for unknown object classes which are not seen during training (epistemic uncertainty).} \Correction{Our definition extends the scope of \emph{robustness} beyond invariance to different perturbations of the input, e.g., 
adversarial attacks~\cite{szegedy2013intriguing, eykholt2017robust}, image corruptions~\cite{pezzementi2018putting, hendrycks2019benchmarking}, change of style~\cite{geirhos2018imagenettrained}, where only prediction accuracy is used as a proxy for robustness. Here, a robust model must not be only accurate, but also well calibrated such that unknown objects or strong input perturbations are designated low confidence scores and easily identified as unreliable and discarded. Although difficult to attain, we argue that both accuracy and calibration are essential for deployment in real world conditions where the data distribution is not identical to the training distribution.\footnote{\Correction{The two metrics are often at odds with each other: a classifier can be accurate but non-calibrated (usually overcofident~\cite{nguyen2015deep, guo2017calibration, hein2019relu}) and conversely it can be inaccurate yet calibrated, if its predictions are always low-confident.}}} To improve the \Correction{robustness} of a DCNN,
given 
the noisy nature of the data, and to address the problem of limited numbers of training labeled images, we propose a technique that combines the teacher-student framework~\cite{mean_teacher_NIPS_2017} with a novel data augmentation strategy (Fig.~\ref{fig:unsupervised_training}). Our augmentation method, named \emph{Superpixel-mix}, exchanges superpixels between training images to generate more training samples that preserve object boundaries and disentangle objects' parts from their frequent contexts. 
To the best of our knowledge, this is among the first investigations on the use of these techniques for reducing DCNN bias and uncertainty.

\noindent \textbf{Contributions: } In summary, our contributions are as follows: \textbf{(i)} Superpixel-mix, a new type of data augmentation for creating new unlabeled images to increase DCNNs' accuracy and \Correction{robustness}.
\textbf{(ii)} A theoretical grounding on why 
mixing augmentation combined with the teacher-student framework can improve \Correction{robustness}. The theory is confirmed by a set of experiments. 
\textbf{(iii)} A new dataset for quantifying contextual bias of DCNNs.\footnote{The dataset will be made publicly available after the anonymity period}

\vspace{-3mm}
\section{Related Work}

\Correction{\noindent\textbf{Robust Deep Learning.} Robustness of DCNNs has been studied under different perspectives in the past few years, e.g., robustness to adversarial attacks~\cite{papernot2016practical, eykholt2017robust}}. We focus here rather on the \Correction{robustness} of the perception functions,
and less on security aspects. Geirhos \etal~\cite{geirhos2018imagenettrained} observe that classification models trained on ImageNet are biased towards textures and blind to shapes. They mitigate this by augmenting the training set with stylized images~\cite{gatys2015neural}, yet this can be detrimental for semantic segmentation as object boundaries are distorted. Shetty \etal~\cite{car_sidewalk_dependency_cvpr_2019} counter contextual bias 
with a data augmentation strategy that removes random objects from images. Some other works focus on evaluating the robustness under different image perturbations, e.g., blur~\cite{vasiljevic2016examining}, brightness~\cite{pei2017deepxplore}. A more systematic study of robustness of classification DCNNs to image perturbations over varying levels of corruption is proposed in \cite{hendrycks2019benchmarking}. This idea is extended to autonomous driving datasets~\cite{michaelis2019benchmarking}, where robustness of object detection methods is evaluated. New datasets with challenging weather conditions, e.g., rain~\cite{hu2019depth}, fog~\cite{sakaridis2018semantic}, low light~\cite{sakaridis2019lowlight} are created to assess and improve robustness of visual perception models. Most approaches simply evaluate evolution of accuracy under such distribution shifts, but ignore other metrics, e.g., calibration that is essential for \Correction{robustness} (calibrated predictions facilitate thresholding for low-confidence predictions and detection of distribution shift). We evaluate our proposed method on multiple shifted datasets~\cite{hendrycks2019benchmarking, michaelis2019benchmarking, sakaridis2018semantic, hu2019depth} and show \Correction{robustness} improvements, beyond accuracy.

\noindent\textbf{DCNN Uncertainty Estimation.} Uncertainty estimation, i.e., knowing when a model does not ``know'' the answer, is a crucial functionality for \Correction{robust} DCNNs. Most DCNN approaches for uncertainty estimation are inspired from Bayesian Neural Networks~\cite{neal1995bayesian, mackay1992bayesian}. Deep Ensembles (DE)~\cite{lakshminarayanan2017simple} train multiple instances of a DCNN with different random initializations, while MC-Dropout~\cite{gal2016dropout} mimics an ensemble through multiple forward passes 
with active Dropout~\cite{srivastava2014dropout} layers. In-between them, some methods generate ensembles with lower training cost (by analyzing weight trajectories during optimization~\cite{maddox2019simple, franchi2019tradi}) or with lower forward cost (by generating ensembles from lower dimensional weights~\cite{wen2020batchensemble, franchi2020encoding} or multiple network heads~\cite{lee2015m}). Other works prioritize computational efficiency to compute uncertainties from a single forward pass~\cite{malinin2018predictive, sensoy2018evidential, postels2019sampling, brosse2020last, van2020uncertainty, joo2020being}, but become specialized to a single type of uncertainty~\cite{hora1996aleatory, kendall2017uncertainty, malinin2018predictive}, e.g., \Correction{Out-Of-Distribution (OOD)}. DE methods are top-performers across benchmarks~\cite{ovadia2019can, gustafsson2020evaluating}. 
Yet, their computational costs make them unfeasible for complex vision tasks, e.g., semantic segmentation. With Superpixel-mix we aim to attain most properties of ensembles, e.g., predictive uncertainty and calibration, in a cost-effective manner.

\noindent\textbf{Augmentation by mixing samples.} Initially seen as a mere heuristic to address over-fitting, data augmentation is now an essential part of recent supervised~\cite{cutmix_ICCV_2019, cutout_arXiv_2017, zhang2017mixup}, semi-supervised~\cite{cutmix_SSL_seg_BMVC_2020, sohn2020fixmatch,mixmatch_nips_2019} and self-supervised learning methods~\cite{gidaris2020learning, chen2020simple}. Mixing techniques, among the most powerful augmentation strategies, generate new ``virtual'' samples (and labels) from pairs of training samples. Mixup\cite{zhang2017mixup} interpolates two images, while Manifold Mixup~\cite{verma2019manifold} interpolates 
hidden activations. CutMix~\cite{cutmix_ICCV_2019,cutmix_SSL_seg_BMVC_2020} replace a random square inside an image with a patch from another image.  Classmix~\cite{classmix_arXiv_2020} cuts and mixes object classes. Puzzle-Mix ~\cite{kim2020puzzle} and Co-Mixup ~\cite{kim2021co}  mix salient areas. For semantic segmentation, mixing by square blocks 
\Correction{is agnostic to} object boundaries and is likely to increase contextual bias as objects \Correction{or object parts can be recognized via their context, i.e., learning shortcuts~\cite{geirhos2020shortcut}.}
Superpixel-mix mitigates this by mixing within object boundaries.

\vspace{-3mm}
\section{Proposed method}

\vspace{-2mm}
\subsection{Overview}\label{Overview}

\begin{figure*}
\centering
\includegraphics[width=0.7\linewidth]{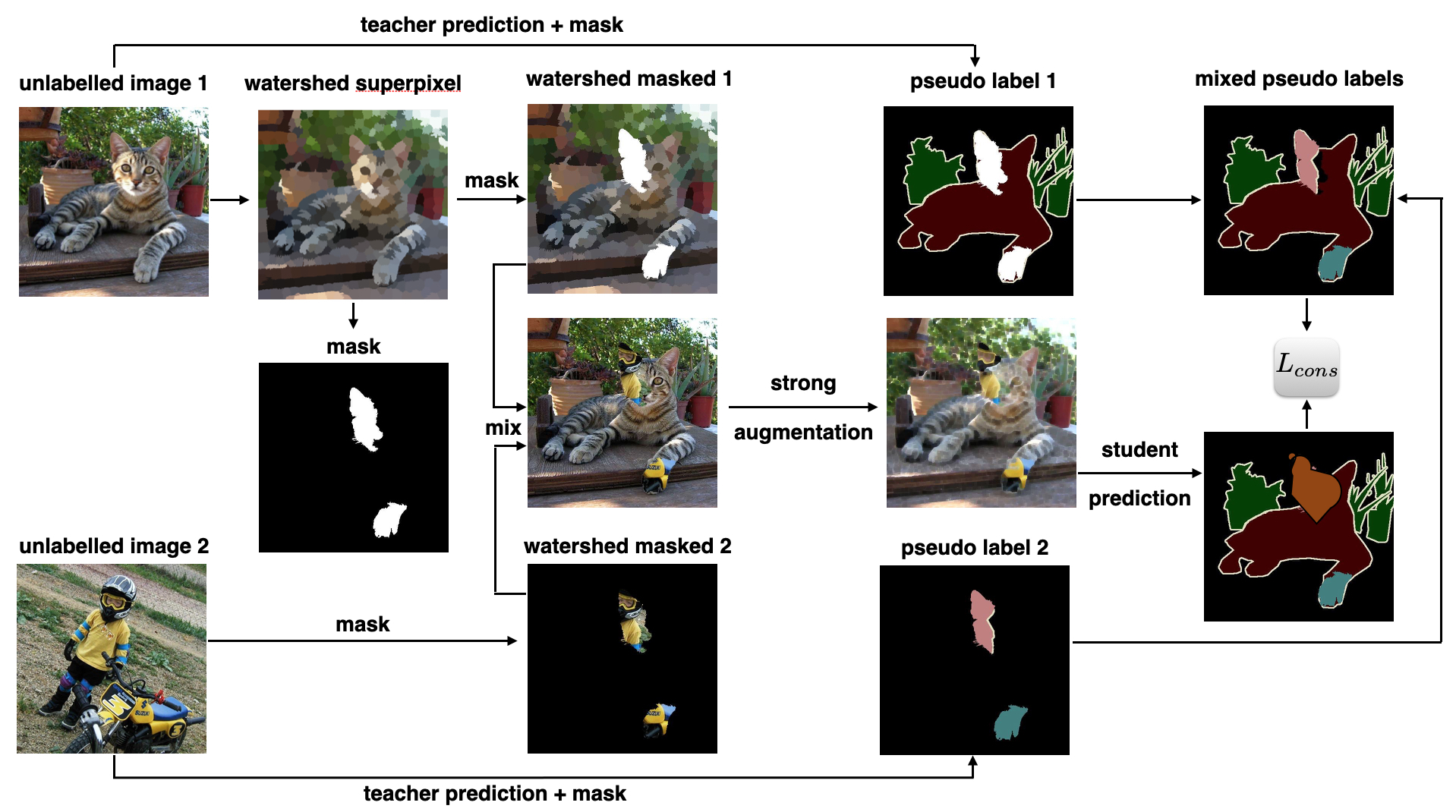}
\vspace{-5pt}
\caption{\small Consistency training with a student-teacher framework using unlabeled images (Task 2). To mix two unlabeled images, superpixels are randomly sampled to create a mixing mask. This mask is used to merge the two images and their pseudo-label outputs from the teacher network. A cross-entropy loss is applied to the student network to encourage consistency between the mixed pseudo-labels and the student network labels.}
\label{fig:unsupervised_training}
\vspace{-10pt}
\end{figure*}

In this paper, we propose a novel superpixel-mix method to generate new training data and 
leverage the results of this data augmentation technique in an existing teacher-student framework~\cite{mean_teacher_NIPS_2017}. %
The combination of our mixing technique and the teacher-student framework forms an optimization component that serves as a consistency constraint in our DCNN training system for semantic segmentation. We conduct experiments for evaluating our trained DCNNs on both types of uncertainties: epistemic and aleatoric. At the end, we also assess our proposed approach in semi-supervised learning.

In general, our training process for all of our experiments comprises two steps that are optimized simultaneously: \textbf{(1)} supervised learning where we train the DCNNs using images with ground-truth labels, and \textbf{(2)} using teacher-student optimization with superpixel-mix data augmentation on images as a consistency constraint that does not use ground-truth labels. In the uncertainty experiments such as \Correction{OOD} in 
\S~\ref{OOD}, DCNNs' bias studying in 
\S~\ref{ablation_bias}, and aleatoric uncertainty in 
\S~\ref{aleatoric}, we use   datasets that contain ground-truth labels for all images. We first train the DCNNs in fully supervised learning fashion as in step 1. We then remove all those labels and use only the images to optimize for the consistency constraint as in step 2. In the semi-supervised learning experiment in 
\S~\ref{semi-supervised}, we use the training dataset that consists of two parts: labelled images and unlabelled images. We train the DCNN using the labelled data for step 1 and the unlabelled data for step 2. %

\textbf{Step 1 - supervised learning with labelled images:}  We use a standard pixel-wise cross-entropy loss denoted by $\mathcal{L}_{\text{sup}}$ and apply 
it to all the labelled images. 
In addition, we use a weak data augmentation (WDA) that consists of \textit{horizontal flipping} and/or \textit{random cropping}.

\textbf{Step 2 - consistency constraint with unlabelled images:} We apply two transformations on an unlabelled image: one is WDA and the other is a strong data augmentation (SDA). Consistency training encourages predictions of the DCNNs to be consistent in the results of the two transformations.  For SDA, we merge WDA and a superpixel mixing technique (see 
\S~\ref{watershedmix}). The consistency loss for optimizing this constraint is denoted as $\mathcal{L}_{\text{cons}}$.

For 
every experiment, we optimize the loss in the overall framework as %
the \textbf{joint loss}  $\mathcal{L} = \mathcal{L}_{\text{sup}}+ \lambda \cdot \mathcal{L}_{\text{cons}}$, where $\lambda$ is a weighting hyper-parameter and is set to 1.

\vspace{-2mm}
\subsection{Teacher-Student Framework}\label{sec:teacherstudent}

To learn from unlabeled images, we follow the teacher-student framework established in~\cite{mean_teacher_NIPS_2017}, where the teacher network produces pseudo-labels learned from the labeled data and the student network is encouraged to be consistent with the teacher.  Consistency is encouraged via 
cross-entropy loss between the two outputs. In our case, we encourage consistency between the mixed output labels of a teacher network 
corresponding to the two unlabelled inputs and the output label of a student network 
predicted from the input that resulted by mixing the two unlabelled images. 
We explain this approach in details in the following paragraph.%

Let $g_{\phi}$ represent the teacher network with weights $\phi$ and $f_{\theta}$ be the student network with weights $\theta$.  For two unlabeled images $x^1_u$ and $x^2_u$, we can use the teacher network to generate pseudo-labels $y^1_u$ and $y^2_u$:
$y^1_u = g_{\phi}(x^1_u)$ and $y^2_u = g_{\phi}(x^2_u)$. Assume now we are given some mixing function $\texttt{mix}$ with a mixing parameter $m$.  Without any assumption on the mixing itself, we denote a mixed output label $y_m$, where $y_m = \texttt{mix}(y^1_u,y^2_u,m)$. For the same mixing parameter $m$, we can also mix the inputs $x^1_u$ and $x^2_u$, \ie $x_m = \texttt{mix}(x^1_u,x^2_u,m)$.
Applying $x_m$ %
to the student network, we expect $f_{\theta}(x_m)$ to be the same as the mixed output $y_m$. This {is} enforced by minimizing the consistency loss $    \mathcal{L}_{\text{cons}} = \text{CE}\left(y_m, %
    f_{\theta}\left(\texttt{mix}(x^1_u, x^2_u, m)\right)\right)$,
where $\text{CE}$ is the pixel-wise cross-entropy and $y_m$ is the mixed pseudo-labels from the teacher.

During training, both the teacher and student networks evolve together. Similarly to~\cite{mean_teacher_NIPS_2017}, we update the teacher network weights $\phi$ after each iteration with an \Correction{Exponential} Moving Average \Correction{(EMA)}, \ie $\phi = \alpha \phi + (1 - \alpha)\theta$ where $\alpha=0.99$ is a momentum-like parameter.%

\vspace{-2mm}
\subsection{Superpixel-mix for semantic segmentation} \label{watershedmix}

To mix two unlabeled images, we use masks generated by sampling superpixels. Superpixels are local clusters of visually similar pixels. Therefore, a group of pixels belonging to the same superpixel are likely to be in the same object or the same part of the object.
There are several superpixel variants, including SEEDS~\cite{van2012seeds}, SLIC~\cite{slic} or Watershed superpixels~\cite{watershed_superpixel_ICIP_2014}.  We opt to use Watershed superpixels as their boundaries retain more salient object edges~\cite{machairas2015waterpixels}.  We refer the reader to the Supplementary Material for details on how we use the watershed transformation to produce superpixels.

Given an unlabeled image $x^1_u$, we apply the watershed superpixel algorithm, which results in a set of $n$ superpixels $\mathcal{S}=\{S_1, S_2,..., S_n\}$.  A mixing mask $m$ (which is a binary mask)\footnote{For simplicity, we overload the notation of the mixing parameter $m$ simply as the mixing mask itself.} is created from a sampled subset of superpixels $S$: $m = \cup_{j\in \sigma(k,n)} S_j$ %
where $\sigma(k,n)$ is a subset of size $k$ of the  $n$  indices, and $k$ is the number of superpixels we want to keep.  %

The mixing mask $m$ defines the pixels in $x^1_u$ which will be replaced by pixels from the unlabeled image $x^2_u$ to form the mixed input $x_m$, \ie $
    x_m = \texttt{mix}(x^1_u, x^2_u, m) = (1-m) \odot x^1_u + m \odot x^2_u
$, where $\odot$ is a pixel-wise multiplication. Superpixels are uniformly sampled given a fixed proportion of selected superpixels. 
Contrary to existing regularization techniques such as Cutout~\cite{cutout_arXiv_2017} or Cutmix~\cite{cutmix_SSL_seg_BMVC_2020}, our superpixel mixing strategy enforces each set of selected pixels in the unlabeled image $x^1_u$ %
to belong to the same object. However, the algorithm is allowed to select a set of superpixel clusters from different objects as illustrated in Figure \ref{fig:unsupervised_training}.  %

We use 200 superpixels per image for all the evaluated datasets. Studies on the number and proportions of superpixels used in mixing are shown in Section~\ref{Ablations_sp} and the Supplement. %

\vspace{-2mm}
\subsection{From empirical risk to teacher-student mixup}

In this section, we show that the training loss of the teacher-student framework in combination with superpixel-mix data augmentation is bounded by the accuracy of the teacher network and the quality of the data augmentation.
Let $\mathcal{D} =\{(x_i,y_i)\}_i \sim \mathcal{P}$ be the labelled dataset which follows the joint distribution $\mathcal{P}$ and $l$ be a loss %
between the target $y$ and the prediction $f_{\theta}(x)$ of the DCNN $f_{\theta}$. Typically, in deep learning, the objective is to learn $\theta$ that minimizes the expected risk defined by: 
$\mathbf{R}_{\mathcal{P}}(f_{\theta}) =\int l(f_{\theta}(x),y)d\mathcal{P}(x,y)$.
As we do not have access to the distribution $\mathcal{P}$, we optimize the loss function that is formed by the empirical risk on $\mathcal{D}$: %
\begin{equation}
    \hat{\mathbf{R}}_{\mathcal{P}_{\delta}}(f_{\theta}) =\frac{1}{n}\sum_{i=1}^n l(f_{\theta}(x_i),y_i) = \int l(f_{\theta}(x),y)d\mathcal{P}_{\delta}(x,y),
\end{equation}
where the summation is converted back to the integral based on $\mathcal{P}_{\delta}(x,y) = \frac{1}{n} \sum_{i=1}\delta(x=x_i,y=y_i)$, as shown by~\cite{zhang2017mixup}.  %

Therefore, we optimize the parameters of the DCNN using the empirical risk. However, the representation of the discretized data is likely to be sparse, Zhang \etal~\cite{zhang2017mixup} proposed to work with $\mathcal{D}_{\text{mix}} =\{(x_{m,i},y_{m,i})\}_i \sim \mathcal{P}^{\text{mix}}_{X,Y}$ where $x_{m,i}$, and $y_{m,i}$ are the data of $\mathcal{D}$ where a mixing procedure has been applied. The hypothesis in~\cite{zhang2017mixup}  is that the mixing procedure helps to better approximate the dataset distribution. Let $ \mathcal{P}^{\text{mix}}_{\delta}$ denote the discrete distribution of this augmented %
dataset. The risk to fit the teacher prediction on $\mathcal{P}^{\text{mix}}_{\delta}$ can then be defined as:
$
\hat{\mathbf{R}}_{\mathcal{P}^{\text{mix}}_{\delta}}(f_{\theta},g_{\phi}) =\int l(f_{\theta}(x),g_{\phi}(x))d\mathcal{P}^{\text{mix}}_{\delta}(x,y).
$ Therefore, our training loss for the overall framework is defined in detail as the following:
$
    \mathcal{L}(\theta)= \hat{\mathbf{R}}_{\mathcal{P}_{\delta}}(f_{\theta})+  \hat{\mathbf{R}}_{\mathcal{P}^{\text{mix}}_{\delta}}(f_{\theta},g_{\phi}).
$ As the loss $l$ is a norm that satisfies the triangle equality, we can prove that the training loss $\mathcal{L}(\theta)$ is bounded by the following:
\begin{equation}
    \mathcal{L}(\theta)\leq 2\mathbf{R}_{\mathcal{P}}(f_{\theta})+  M(\| \mathcal{P}^{\text{mix}}_{\delta} -\mathcal{P}\|_1 + \| \mathcal{P}_{\delta}  -\mathcal{P}\|_1) +\hat{\mathbf{R}}_{\mathcal{P}^{\text{mix}}_{\delta}}(g_{\phi}),
\end{equation}

\noindent where the four terms are linked to the true error, approximation error, mixing distribution error and teacher error, respectively. 
This implies that the quality of the DCNN is bounded by the accuracy of the teacher. It is also bounded by how much the mixing strategy can sample the true distribution of the dataset. Finally, the distribution of the training data with respect to the true data distribution also plays an important role.  %
This finding can be applied to all teacher-student frameworks, such as those used in SSL, self supervised training, and domain adaptation.
The detailed proof and analysis is given in the Supplementary Material. %

\Correction{$\| \mathcal{P}^{\text{mix}}_{\delta} -\mathcal{P}\|_1$ and $\| \mathcal{P}_{\delta}  -\mathcal{P}\|_1$
reflect the capacity of the two  distributions $\mathcal{P}_{\delta}$ and $\mathcal{P}^{\text{mix}}_{\delta}$ to approximate the true dataset distribution.  This bound allows us to control the variation of the risk. To reduce the risk, we can increase the number of training data or improve the quality of the data augmentation and so  reduce $\| \mathcal{P}^{\text{mix}}_{\delta} -\mathcal{P}\|_1 $. This motivates our research on data augmentation strategies to approximate the true distribution in a data-efficient way. We can also improve the quality of the teacher using EMA training that stabilizes the training loss. } %

\vspace{-3mm}
\subsection{Uncertainty and Deep Learning}

Consider a joint distribution $\mathcal{P}$ over input $x$ and labels $y$ over a set of labels $\mathcal{Y}$. When a DCNN performs inference, it predicts $f_{\theta} = \mathcal{P}(y|x,\theta)$, where $\theta$ is optimized to minimize the loss over the training set $\mathcal{D}$. This likelihood typically suffers from two kinds of uncertainty\AB{~\cite{hora1996aleatory, kendall2017uncertainty}}.  First is aleatoric uncertainty, linked to the unpredictability of the data acquisition process. During inference, instead of working with $x$, we may have access to $x+n$, %
where $n$ represents noise on the input data. Second is epistemic uncertainty, linked to the lack of knowledge of the model, \AB{i.e.,} the weights $\theta$ of the network. In addition, the epistemic uncertainty can be subdivided into two sub-types: one linked to the \Correction{OOD} \AB{~\cite{malinin2018predictive}} and the other one linked to networks' bias. \Correction{Epistemic uncertainty models the uncertainty associated with limited sizes of training datasets. Most works focus only on the ability to detect OOD. In this paper, we conduct experiments for all types of uncertainties: aleatoric (i.e., testing models on noisy data) and two sub-types of epistemic (i.e., OOD detection and models' bias).}%

\vspace{-3mm}
\section{Experiments}

We 
\AB{conduct} experiments on five datasets. First, %
we study 
\AB{network robustness} to epistemic uncertainty experiments on StreetHazards \cite{hendrycks2019anomalyseg}.  
\AB{The test set contains some object classes that are not available in the training set.}
The goal is to detect these out-of-distribution (OOD) classes. We also evaluate the performance of the DCNNs on an contextually unbiased dataset.
Furthermore, we investigate the networks' \Correction{robustness} for the aleatoric uncertainty. 
\AB{To this end, we train a DCNN on Cityscapes ~\cite{Cityscapes_CVPR_2016}} and evaluate the performances on Rainy \cite{hu2019depth} and Foggy Cityscapes~\cite{sakaridis2018semantic}. %
Finally, we evaluate \AB{on the} semi-supervised learning \AB{task} on Cityscapes~\cite{Cityscapes_CVPR_2016} and Pascal ~\cite{pascal-voc-2012}. We implement the experiments using PyTorch (see Supplementary Material).

\vspace{-3mm}
\subsection{Evaluation criteria}

The first criterion we use is the mIoU \cite{jaccard1912distribution}, 
\AB{reflecting the predictive performance of segmentation models.}
\AB{Second, similarly to \cite{lakshminarayanan2017simple} we use} the negative log-likelihood (NLL), \AB{a proper scoring rule~\cite{gneiting2007strictly},} which depends on the aleatoric uncertainty \Correction{and can assess the degree of overfitting~\cite{guo2017calibration}.}. In addition, we use the expected calibration error (ECE) ~\cite{guo2017calibration} that measures 
\AB{how} the confidence score predicted by a DCNN is related to its accuracy. Finally, we use the AUPR , AUC, and the FPR-95-TPR  defined in \cite{hendrycks2016baseline} that evaluate the quality of a DCNN to detected OOD data. With multiple metrics, we can get a clearer picture on the performance of the DCNNs with regards to accuracy, calibration error, failure rate, \AB{OOD detection}. Even though it is difficult to achieve top performance on all metrics, 
\AB{we argue that it is more pragmatic and convincing}
to evaluate on multiple metrics~\cite{ovadia2019can, fort2019deep} than \AB{optimizing for a single metric, potentially at the expense of many others.}
\Correction{For example, a DCNN with a low accuracy and a low confidence score is well-calibrated. Therefore, evaluating a DCNN on a single metric such as ECE or mIoU alone is not enough. We aim to have a good compromise between accuracy and calibration.}

 \vspace{-3mm}
\subsection{Epistemic uncertainty: Out-Of-Distribution (OOD) Detection }\label{OOD}

\Correction{One cause of the epistemic uncertainty in deep learning is the limited training data that does not cover all possible object classes. The evaluation of this type of epistemic uncertainty is often linked to OOD detection. Therefore, this experiment is designed to evaluate the epistemic uncertainty using StreetHazards~\cite{hendrycks2019anomalyseg}.}
StreetHazards is a large-scale dataset that consists of different sets of synthetic images of street scenes. This dataset is composed of $5,125$ images for training and $1,500$ test images.
The training dataset contains pixel-wise annotations for $13$ classes. The test dataset comprises $13$ training classes and $250$ OOD classes, unseen in the training set, making it possible to test the robustness of the algorithm when facing a diversity of possible scenarios.
For this experiment, we use DeepLabv3+~\cite{chen2018encoder} with the experimental protocol from~\cite{hendrycks2019anomalyseg}. We use ResNet50 encoder~\cite{he2016deep}. For this experiment, we compare our algorithm to Deep Ensembles \cite{lakshminarayanan2017simple}, BatchEnsemble \cite{wen2020batchensemble}, LP-BNN \cite{franchi2020encoding}, TRADI \cite{franchi2019tradi} , MIMO \cite{havasi2020training}
\AB{achieving} state-of-the-art results on epistemic uncertainty. We also compare our \AB{model with} MCP which is the baseline DCNN where we consider the maximum probability class as a confidence score, and with Cutmix~\cite{cutmix_SSL_seg_BMVC_2020} strategy.
The results in Table~\ref{table:outofditribution} show that our method is the only one to have the best results in three out of five measures. Running up, Deep Ensemble and LP-BNN achieve best results on one measure only. Moreover, our method achieves the best results faster than Deep Ensemble and LP-BNN, we only need one inference pass compared to 4 inference passes for the others.

\begin{table*}[!t]
\renewcommand{\figurename}{Table}
 \begin{center}
 \scalebox{0.60}
 {
 \begin{tabular}{c l  c c c c c c }
 \toprule
 Dataset & OOD method  & mIoU $\uparrow$ & AUC $\uparrow$  & AUPR $\uparrow$ & FPR-95-TPR $\downarrow$ & ECE $\downarrow$ & \# Forward passes $\downarrow$ \\ 
 \midrule
\multirow{6}{*}{\shortstack[c]{\textbf{StreetHazards} \\ DeepLabv3+ \\ ResNet50}}  &
Baseline  (MCP)~\cite{hendrycks2016baseline} & 53.90\%  & 0.8660& 0.0691 & 0.3574 & 0.0652 &1 \\ 
             & TRADI      \cite{franchi2019tradi} &  52.46\%	& 0.8739 & 0.0693 & 0.3826 &0.0633 &4 \\ 
            & Cutmix ~\cite{cutmix_SSL_seg_BMVC_2020}  & 56.06\% & 0.8764 & 0.0770 &0.3236 &0.0592 &1 \\ 
            & MIMO  \cite{havasi2020training} & 55.44\% &  0.8738 &  0.0690 & 0.3266 & 0.0557 &4 \\ 
            & BatchEnsemble \cite{wen2020batchensemble}  &56.16\%  & 0.8817 & 0.0759 & 0.3285 & 0.0609 &4 \\ 
      & LP-BNN \cite{franchi2020encoding}  &  54.50\% & 0.8833 & 0.0718 & 0.3261 & \textbf{0.0520} &4 \\ 
    & Deep Ensembles  \cite{lakshminarayanan2017simple}& 55.59\% & 0.8794 & \textbf{0.0832} & 0.3029 & 0.0533  &4 \\ 
   & Superpixel-mix (ours) & \textbf{56.39}\% & \textbf{0.8891} & 0.0778 & \textbf{0.2962} &0.0567 &1 \\ 

\bottomrule
 \end{tabular}
 } %
 \end{center}
 \vspace{-1mm}
 \caption{\small \textbf{Comparative results on the OOD task for semantic segmentation.}  \small Results are averaged over three seeds.}\label{table:outofditribution}
 \vspace{-3mm}
 \end{table*}

 \vspace{-3mm}
\subsection{Epistemic uncertainty : Unbiased experiment} \label{ablation_bias}
\vspace{-1mm}

\begin{figure*}[!t]
\centering
\includegraphics[width=0.5\linewidth]{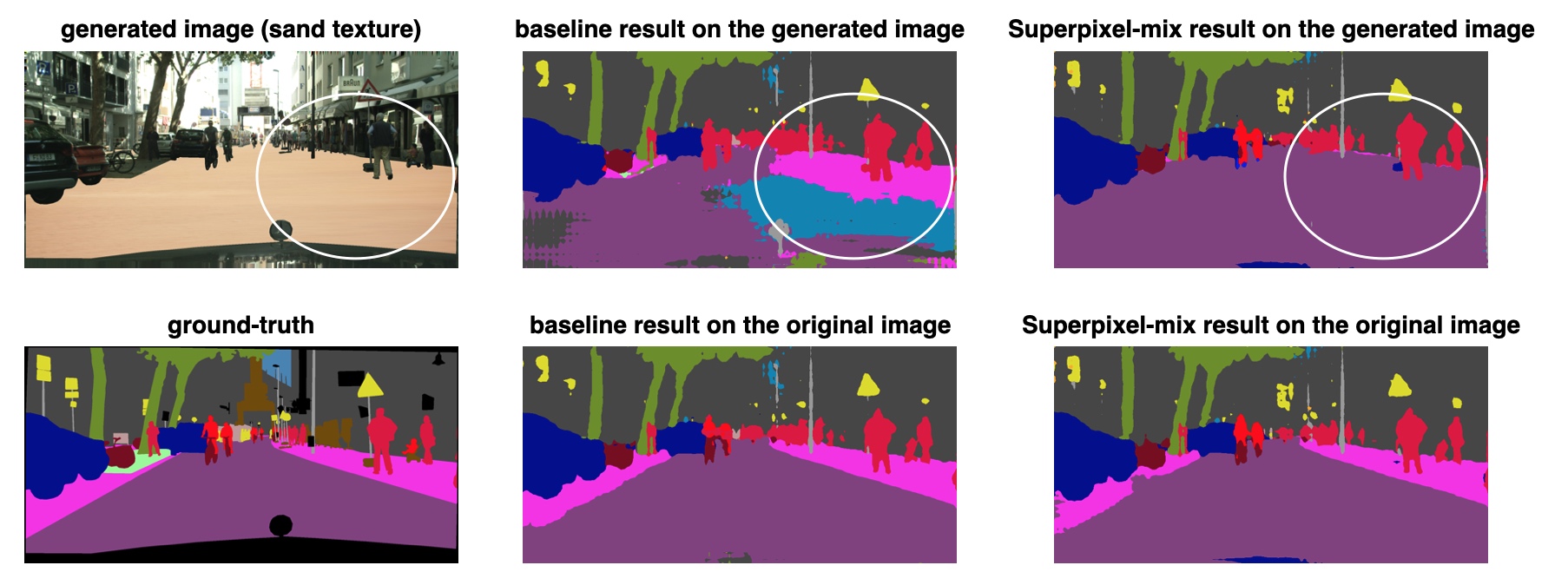}
\vspace{-3mm}
\caption{\small A qualitative example of the network bias study. When the road and pavements are replaced with sand texture, the baseline supervised network makes wrong segmentations. There is still pavement segmentation due to the association with people. Superpixel-mix produces better results without the wrong pavement segmentation, also provides a clean segmentation for the sand texture.}
\label{fig:bias_study}
\vspace{-5mm}
\end{figure*}

\Correction{The second aspect of epistemic uncertainty 
is related to network biased caused limited samples and diversity of scenarios, e.g., certain objects always co-occur in the training data. 
Here, we evaluate the epistemic uncertainty under the lens of model bias.
}
In urban datasets such as Cityscapes, 
\Correction{road and car pixels appear most of the time together,} raising the question of co-occurrence dependency between objects. If two objects are dependent, then the network will likely fail when the car object is encountered in different context other than roads. Shetty \etal~\cite{car_sidewalk_dependency_cvpr_2019} study object dependency and suggest using GANs to remove one object by in-painting and training the network for the new contexts. Their results show that a network that is less biased towards the co-occurrence dependency yields better accuracy for segmentation.

To measure the bias of our superpixel-mix method, we create a new dataset dubbed 
 Out-of-Context Cityscapes (OC-Cityscapes), by replacing roads in the validation data of Cityscapes with various textures such as water, sand, grass, etc. Example images are shown in the Supplementary Material. Studies in~\cite{geirhos2018imagenettrained} show that DCNNs are biased towards texture. By replacing different textures for roads, we test the trained networks on these new context images and evaluate the bias level for each network.

In Table~\ref{table:bias_study}, we show the performances of the fully supervised network (baseline) and our network trained using the superpixel-mix method for the Cityscapes validation set and our generated dataset. The results are mIoU, ECE, and NLL scores averaged over 3 runs for classes that are not road. On our experimental dataset, 
the baseline's performance drops by 21.97\% while 
Superpixel-mix drops by 19.83\% for the mIoU metric. The results also show that 
Superpixel-mix 
produces a less biased model for 
co-occurring objects with the best 
scores on mIoU and NLL measures. %
Superpixel-mix changes the image context while preserving object shapes, effectively regularizing the model to address shortcut learning, i.e., overfitting on the co-occurrence of objects and their typical contexts. 
For a visual example, see Figure~\ref{fig:bias_study}.

\begin{table}[!t]
\renewcommand{\figurename}{Table}
\begin{center}
 \scalebox{0.50}
 {
\begin{tabular}{ l  c  c  c  c  c c  }
\toprule
Evaluation data & Cityscapes  & OC-Cityscapes &Cityscapes  & OC-Cityscapes & Cityscapes  & OC-Cityscapes \\
    &  mIoU &  mIoU &  NLL & NLL & ECE  & ECE \\
\midrule
Baseline (MCP)~\cite{hendrycks2016baseline} & 76.51\% & 54.54\% & -0.9456 & -0.7565 & 0.1303 & 0.2162 \\
 Cutmix ~\cite{cutmix_SSL_seg_BMVC_2020} & 78.37\% & 54.78\%  & -0.955&	-0.7435 & 0.1365 & 0.2587\\
 MIMO  \cite{havasi2020training}  & 77.13\% & 55.87\% & -0.9516 & -0.7431 & 0.1398 & 0.2359\\
Deep Ensembles  \cite{lakshminarayanan2017simple} & 77.48\% & 57.09\%  & -0.9469 & -0.7613 & \textbf{0.1274} & \textbf{0.1968}\\
Superpixel-mix (ours) & \textbf{78.99\%} & \textbf{59.16\%} & \textbf{-0.9563} & \textbf{-0.7768} & 0.1348 & 0.2244\\
\bottomrule
\end{tabular}
}
\end{center}
\vspace{-1mm}
\caption{\small \textbf{Comparative results for network biases on OC-Cityscapes.} The results are segmentation mIoU, NLL and ECE, for classes that are not road. The baseline is the result from supervised training.}
\label{table:bias_study}
\vspace{-2mm}
\end{table}

\vspace{-3mm}
\subsection{Aleatoric uncertainty experiments}\label{aleatoric}

\Correction{Aleatoric uncertainty is associated with unpredictability of the data acquisition process that causes various noises in the data.
In the following experiments we evaluate the aleatoric uncertainty of DCNNs trained on normal images (e.g., normal weather images) when facing test images with various types of noise (e.g., rainy or foggy environments).}
In semantic segmentation, the DCNN must be \Correction{robust} to aleatoric uncertainty. To check that, we use the rainy \cite{hu2019depth} and foggy Cityscapes \cite{sakaridis2018semantic} datasets, which are built by adding rain of fog to the Cityscapes validation \Correction{images}. The goal 
is to evaluate the performance of DNNs to resist these perturbations. 
\Correction{We further generate an additional Cityscapes variant with images modified with different perturbations and intensities to mimic a distribution shift~\cite{hendrycks2019benchmarking}.} 
\Correction{We apply the following} perturbations: Gaussian noise, shot noise, impulse noise, defocus blur, frosted, glass blur, motion blur, zoom blur, snow, frost, fog, brightness, contrast, elastic, pixelate, JPEG. For more information, please refer to \cite{hendrycks2019benchmarking}. We call this dataset Cityscapes-C.

To measure the \Correction{robustness} under aleatoric uncertainty, we 
compute ECE, mIoU and NLL scores averaged over 3 runs. Table \ref{table:Aleatoric} shows results close to the state of the art. 
DE reaches good results, yet this approach needs to train several DCNNs, so it is more time-consuming for training and inference. In the Supplementary Material, we 
report mIoU scores of different approaches for the different levels of noise. Overall, our 
experiments indicate that Superpixel-mix tends to be 
\Correction{robust} to high level of noise.

\begin{table}[t]
\begin{center}
 \scalebox{0.50}
 {
\begin{tabular}{ l  c  c  c  c c c c c c c c c   }
\toprule
\multirow{2}{*}{Evaluation data} & \multicolumn{3}{c}{Cityscapes} & \multicolumn{3}{c}{Rainy Cityscapes}  & \multicolumn{3}{c}{Foggy Cityscapes}  & \multicolumn{3}{c}{Cityscapes-C}  \\
    &  mIoU $\uparrow$ &  ECE $\downarrow$  &  NLL $\downarrow$  &  mIoU $\uparrow$  &  ECE $\downarrow$ &  NLL $\downarrow$  &  mIoU $\uparrow$ &  ECE $\downarrow$  &  NLL $\downarrow$  & mIoU $\uparrow$    & ECE $\downarrow$ &  NLL $\downarrow$   \\
\midrule
Baseline (MCP)~\cite{hendrycks2016baseline} &  76.51\% & 0.1303 &-0.9456 & 58.98\% & 0.1395  & -0.8123 & 69.89\%   & 0.1493 &-0.9001  & 40.85\%   & 0.2242 &-0.7389 \\
 Cutmix ~\cite{cutmix_SSL_seg_BMVC_2020}  & 78.37\% & 0.1365  & -0.9550& 61.86\% & 	0.1559& -0.8200 &  73.57\%  & 0.1484 &-0.9289 & 39.16\% 	& 0.3064 &	-0.6865\\
MIMO  \cite{havasi2020training}  &  77.13\% & 0.1398  &-0.9516 & 59.27\% & 0.1436 & -0.8135 &70.24\% & 0.1425 &-0.9014& 40.73\%    & 0.2350 &-0.7313\\
BatchEnsemble \cite{wen2020batchensemble} &  77.99\% & 0.1129 &-0.9472 & 60.29\% & 0.1436 &-0.7820& 72.19\%  & 0.1425 &-0.9132&   40.93\% &  0.2270 &-0.7082\\
LP-BNN  \cite{franchi2020encoding}  &  77.39\% & \textbf{0.1105}  &-0.9464 & 60.71\% & 0.1338 &-0.7891& 72.39\% & \textbf{0.1358} &-0.9131& \textbf{43.47}\%    & 0.2085 &-0.7282\\
Deep Ensembles  \cite{lakshminarayanan2017simple} &  77.48\% & 0.1274 &-0.9469 & 59.52\%  &\textbf{0.1078}   &-0.8205& 71.43\% & 0.1407  &-0.9070 & 43.40\%  & \textbf{0.1912}&-0.7509 \\
Superpixel-mix  (ours)  &   \textbf{78.99}\% & 0.1348 & \textbf{-0.9563}  & \textbf{61.87}\% & 	0.1583& 	\textbf{-0.8207}& \textbf{74.39}\%  & 0.1411 &-\textbf{0.9266}& 42.58\%  & 	0.2338 &	\textbf{-0.7513}\\

\bottomrule
\end{tabular}
}
\end{center}
\caption{\small Aleatoric uncertainty study on Cityscapes-C,  Foggy Cityscapes~\cite{hu2019depth}  and Rainy Cityscapes~\cite{sakaridis2018semantic}.  }
\label{table:Aleatoric}
\vspace{-4mm}
\end{table}

\Correction{We note that our method does not achieve the top ECE scores across the various perturbations and weather conditions in these experiments. However, none of the considered strong baselines based on ensembles outperforms the others consistently either due to the difficulty and diversity of the considered test sets. Superpixel-mix achieves competitive ECE scores and is a top performer on mIoU and NLL.}

\vspace{-3mm}
\subsection{Semi-Supervised Learning experiments}\label{semi-supervised}
To evaluate the \Correction{robustness} of our method to missing annotation, we tested our approach on a semi-supervised learning task on two datasets: Cityscapes~\cite{Cityscapes_CVPR_2016} and Pascal VOC 2012~\cite{pascal-voc-2012}. 
We follow the common practice for this task from prior works~\cite{adversarial_bmvc_2018,cutmix_SSL_seg_BMVC_2020} and use DeepLab-V2~\cite{chen2017deeplab} model with ResNet101~\cite{he2016deep} encoder pre-trained on ImageNet~\cite{ImageNet_cvpr_2009} and MS-COCO~\cite{mscoco_eccv_2014}.
The weights of both the teacher and student models are initialized the same manner.
We evaluate our method and compare with existing methods using three sets of labeled data: 1/30 (100 images), 1/8 (372 images) and 1/4 (744 images). Our results are reported as average mIoU of 12 runs (4 times on each of 3 official splits) %
as well as the standard deviation. The results are shown in Table \ref{table:Cityscapes}. We present results on Pascal dataset in the Supplementary Material.

\begin{table*}[!t]
\renewcommand{\figurename}{Table}
\begin{center}
\scalebox{0.55}
 {
\begin{tabular}{l  l l l}
\toprule
\textbf{Labeled samples} & \textbf{1/30 (100)} & \textbf{1/8 (372)} & \textbf{1/4 (744)} \\
\midrule
Adversarial~\cite{adversarial_bmvc_2018} & - & 58.80\% & 62.30\% \\
s4GAN~\cite{s4GAN_pami_2019} & - & 59.30\% &  61.90\% \\
Cutout~\cite{cutout_arXiv_2017} & 47.21\% $\pm$ 1.74 & 57.72\% $\pm$ 0.83 & 61.96\% $\pm$ 0.99 \\
Cutmix~\cite{cutmix_SSL_seg_BMVC_2020} & 51.20\% $\pm$ 2.29 & 60.34\% $\pm$ 1.24 & 63.87\% $\pm$ 0.71 \\
Classmix~\cite{classmix_arXiv_2020} & 54.07\% $\pm$ 1.61 & 61.35\% $\pm$ 0.62 & 63.63\% $\pm$ 0.33 \\
Classmix~\cite{classmix_arXiv_2020} & 54.07\% $\pm$ 1.61 & 61.35\% $\pm$ 0.62 & 63.63\% $\pm$ 0.33 \\
\midrule
Baseline(*) & 43.84\% $\pm$ 0.71 & 54.84\% $\pm$ 1.14 & 60.08\% $\pm$ 0.62\\
Superpixel-mix (ours) & \textbf{54.11\% $\pm$ 2.88  ($\uparrow$ 7.27\%)} &\textbf{63.44\%$\pm$ 0.88  ($\uparrow$ 8.60\%)} & 	\textbf{65.82\%$\pm$ 1.78  ($\uparrow$ 5.74\%) }\\

\bottomrule
\end{tabular}
}
\end{center}
\caption{\small 
\textbf{Evaluation in the semi-supervised learning regime on Cityscapes. We report mIoU scores as $mean \pm std.dev$ computed over 12 runs.}
The $(\uparrow)$ shows the improvement of our methods over the baselines. (*) The baselines are from~\cite{classmix_arXiv_2020} as we use a similar base procedure.}
\label{table:Cityscapes}
\end{table*}

\vspace{-3mm}
\subsection{Ablations and analysis}\label{Ablations_sp}
\Correction{We perform various ablations to understand the 
influence of different hyper-parameters and choices
in our algorithm. First, we vary the number of superpixels extracted per image from 20 to 1,000. The results (in Table 5, Supplement) show that our method achieves the highest mIoU on Cityscapes when the number of superpixels per image is from 100 to 200. Secondly, we study the proportion of chosen superpixels for mixing over the total number of superpixels per image. The proportion ranges from 0.1 to 0.9. We find that the results vary little across all the proportion values (Table 6, Supplement). The best mIoU is obtained at the proportion value of 0.6. Finally, we study the influence of different superpixel techniques to generate mixing masks: Watershed~\cite{watershed_superpixel_ICIP_2014}, SLIC~\cite{slic}, and Felzenszwalb~\cite{felzenswalb2004efficient}. The results in Table~\ref{table:Abblation000} show that on Cityscapes segmentation the performance of Superpixel-mix is relatively stable across superpixel methods, with Watershed yielding the best mIoU score.}

\begin{table}[t]
\begin{center}
 \scalebox{0.55}
 {
\begin{tabular}{l| c  c  c   }
\toprule
superpixels algorithm &  Watershed ~\cite{watershed_superpixel_ICIP_2014} &  SLIC~\cite{slic}  & Felzenszwalb  \cite{felzenswalb2004efficient} \\

 mIoU & 78.99 \%& 78.89\% & 77.99\%  \\

\bottomrule
\end{tabular}
}
\end{center}
\caption{\small \textbf{Ablation study results over influence of different superpixel techniques. We report mIoU scores for semantic segmentation on Cityscapes.}}
\label{table:Abblation000}
\vspace{-5mm}
\end{table}

\vspace{-3mm}
\section{Conclusions}

Superpixel-mix data augmentation is a promising new training technique for semantic segmentation. This strategy for creating diverse data, combined with a teacher-student framework, leads to better accuracy and to more \Correction{robust} DCNNs.
 We are the first, to the best of our knowledge, to successfully apply the watershed algorithm in data augmentation. What sets our data augmentation technique apart from existing methods is the ability to preserve the global structure of images and the shapes of objects while creating image perturbations. We conduct various experiments with different types of uncertainty. The results show that our strategy achieves state-of-the-art \Correction{robustness} scores for epistemic uncertainty. For aleatoric uncertainty, our approach produces state-of-the-art results in Foggy Cityscapes and Rainy Cityscapes. In addition, our method needs just one forward pass.

Previous research in machine learning has established that creating more training data using data augmentation may improve the accuracy of the trained DCNNs substantially. Our work not only confirms that, but also provides evidence that some data augmentation methods, such as our Superpixel-mix, help to improve the \Correction{robustness} of DCNNs by reducing both epistemic and aleatoric uncertainty.

\section*{Acknowledgments}
\Correction{This work was performed using HPC resources
from GENCI-IDRIS (Grant 2020-AD011011970) and (Grant 2021-AD011011970R1).}
\bibliography{egbib}

\clearpage
\begin{widetext}
\begin{center}
\textbf{\large Robust Semantic Segmentation with Superpixel-Mix (Supplementary material)}
\end{center}
\end{widetext}

\section{Watershed tranform for Superpixel-mix} \label{sup_watershedmix}

\begin{figure}[h]
 \centering
\includegraphics[width=0.8\linewidth]{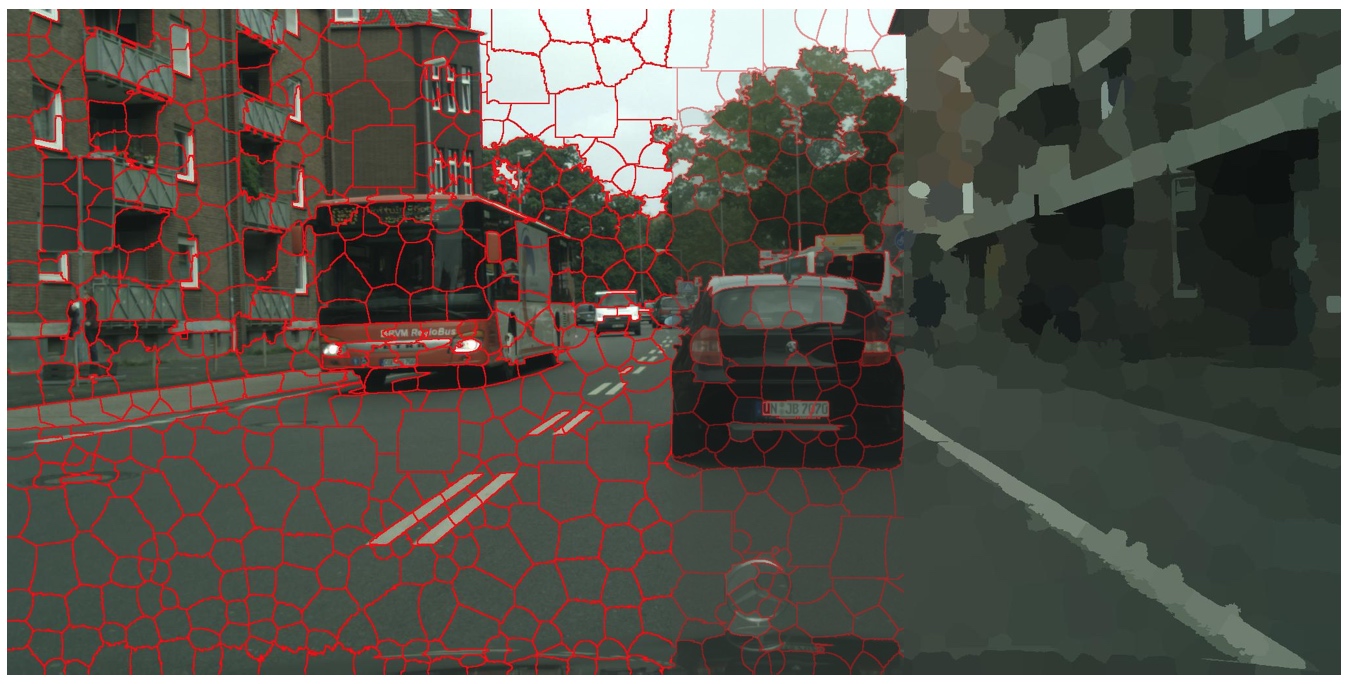}
 \caption{An example of superpixels on  an image from Cityscapes dataset.  } %
 \label{fig:seg-example}
 \end{figure}

To mix two unlabeled images, we use masks generated 
\ab{from randomly sampled} superpixels. Superpixels are local clusters of visually similar pixels\ab{, typically delimited by pronounced edges} {(as illustrated in Figure \ref{fig:seg-example})}. Therefore, a group of pixels belonging to the same superpixel are likely 
\ab{to correspond to the same object or a part of an object.}
\ab{There are various methods for computing superpixels, including SEEDS~\cite{van2012seeds}, SLIC~\cite{slic} or Watershed~\cite{watershed_superpixel_ICIP_2014}}.
We opt to use Watershed superpixels as their boundaries retain more salient object edges~\cite{machairas2015waterpixels}.  

Watershed transformation and all its variants~\cite{beucher1994watershed,beucher1993morphological,beucher1979use} are powerful techniques for image segmentation. 
\ab{Watershed processes image gradients and outputs corresponding clusters for each pixel.}
Since the watershed input is a gradient, we
\ab{convert the input image from RGB to Lab in order to compute the gradient maps.} 
Then, on each channel, we evaluate a morphological gradient \ab{and we average the three results.}
Similarly to~\cite{watershed_superpixel_ICIP_2014}, instead of considering all the clusters of the watershed, we build a regular grid of points and consider these points as markers for the watershed transform. This strategy allows us to control the number of superpixels to reduce computational cost.

\section{From empirical risk to teacher student mixup}

In this section, we show that the training loss of the teacher-student framework in combination with superpixel-mix data augmentation is bounded by the accuracy of the teacher network and the quality of the data augmentation.
{This result is essential since it is the first bound for a teacher-student framework with consistency training. 
\ab{Different and effective variants of teacher-students approaches with consistency training have emerged in recent literature, in particular in semi-supervised learning~\cite{berthelot2019mixmatch, berthelot2019remixmatch, sohn2020fixmatch} and self-supervised learning~\cite{grill2020bootstrap, gidaris2020learning, chen2021exploring}.
All these approaches rely heavily on well-crafted agressive data augmentation strategies. This result could be useful in this context as we proove how the quality of the teacher and the quality of the data augmentation influence the accuracy of the student.
}
}

Let $\mathcal{D} =\{(x_i,y_i)\} \sim \mathcal{P}$ be the labelled dataset which follows the joint distribution $\mathcal{P}$ and $l$ be a loss between the target $y$ and the prediction $f_{\theta}(x)$ of the DCNN $f_{\theta}$. Typically, in deep learning, the objective is to learn $\theta$ that minimizes the expected risk defined by: 
$\mathbf{R}_{\mathcal{P}}(f_{\theta}) =\int l(f_{\theta}(x),y)d\mathcal{P}(x,y)$.
As we do not have access to the distribution $\mathcal{P}$, we optimize the loss function that is formed by the empirical risk on $\mathcal{D}$: %
\begin{equation}
    \hat{\mathbf{R}}_{\mathcal{P}_{\delta}}(f_{\theta}) =\frac{1}{n}\sum_{i=1}^n l(f_{\theta}(x_i),y_i) = \int l(f_{\theta}(x),y)d\mathcal{P}_{\delta}(x,y),
\end{equation}
where the the summation is converted back to the integral based on $\mathcal{P}_{\delta}(x,y) = \frac{1}{n} \sum_{i=1}\delta(x=x_i,y=y_i)$, as shown by~\cite{zhang2017mixup}.  %

Therefore, we optimize the parameters of the DCNN using the empirical risk. 
\ab{However, the available training samples offer only a limited sparse coverage of the data distribution.}
\ab{Aiming to achieve a better and denser coverage of the data distribution, Zhang et al.~\cite{zhang2017mixup} propose working instead with $\mathcal{D}_{\text{mix}} =\{(x_{m,i},y_{m,i})\}_i \sim \mathcal{P}^{\text{mix}}_{X,Y}$ where $x_{m,i}$, and $y_{m,i}$ are obtained from pairs samples from $\mathcal{D}$ mixed together.}
The hypothesis in~\cite{zhang2017mixup} is that the mixing procedure 
\ab{enables a better coverage and, consequently, approximation of the dataset distribution.} Let $ \mathcal{P}^{\text{mix}}_{\delta}$ denote the discrete distribution of this augmented %
dataset. \ab{Zhang et al.~\cite{zhang2017mixup} argue that the naïve estimate $\mathcal{P}_{\delta}$ is merely a suboptimal approximation out of the many possible choices towards approximating the true distribution $\mathcal{P}$. Inspired by the vicinal risk minimization principle~\cite{chapelle2001vicinal} that estimates distributions around data samples, they argue that $\mathcal{P}^{\text{mix}}_{\delta}$ is a better approximation as it covers inter-sample areas through sample mixing, i.e., generating \emph{virtual} samples. Here, we build upon this finding from~\cite{zhang2017mixup} and consider images computed with Superpixel-mix also as virtual samples from the vicinal distribution of the original samples.} 
The {vicinal} risk to fit the teacher prediction on $\mathcal{P}^{\text{mix}}_{\delta}$ can then be defined as:

\begin{equation}
\hat{\mathbf{R}}_{\mathcal{P}^{\text{mix}}_{\delta}}(f_{\theta},g_{\phi}) =\int l(f_{\theta}(x),g_{\phi}(x))d\mathcal{P}^{\text{mix}}_{\delta}(x,y).
\end{equation}

\noindent Therefore, our training loss for the overall framework is defined in detail as the following:

\begin{equation}
    \mathcal{L}(\theta)= \hat{\mathbf{R}}_{\mathcal{P}_{\delta}}(f_{\theta})+  \hat{\mathbf{R}}_{\mathcal{P}^{\text{mix}}_{\delta}}(f_{\theta},g_{\phi}).
\end{equation}

\noindent As the loss $l$ is a norm that satisfies the triangle equality, we can prove that the training loss $\mathcal{L}(\theta)$ is bounded by the following:
\begin{equation}
    \mathcal{L}(\theta)\leq 2\mathbf{R}_{\mathcal{P}}(f_{\theta})+  M(\| \mathcal{P}^{\text{mix}}_{\delta} -\mathcal{P}\|_1 + \| \mathcal{P}_{\delta}  -\mathcal{P}\|_1) +\hat{\mathbf{R}}_{\mathcal{P}^{\text{mix}}_{\delta}}(g_{\phi}),
\end{equation}
\noindent where the four terms are linked to the true error, mixing distribution error, approximation error and finally the teacher error. %

\begin{proof}

\begin{equation}
    \mathcal{L}(\theta)= 2(\mathbf{R}_{\mathcal{P}}(f_{\theta}) - \mathbf{R}_{\mathcal{P}}(f_{\theta}))+\hat{\mathbf{R}}_{\mathcal{P}_{\delta}}(f_{\theta})+  \hat{\mathbf{R}}_{\mathcal{P}^{\text{mix}}_{\delta}}(f_{\theta},g_{\phi}) + \hat{\mathbf{R}}_{\mathcal{P}^{\text{mix}}_{\delta}}(f_{\theta})- \hat{\mathbf{R}}_{\mathcal{P}^{\text{mix}}_{\delta}}(f_{\theta})
\end{equation}

\begin{equation}
    \mathcal{L}(\theta)\leq |2(\mathbf{R}_{\mathcal{P}}(f_{\theta}) - \mathbf{R}_{\mathcal{P}}(f_{\theta}))+\hat{\mathbf{R}}_{\mathcal{P}_{\delta}}(f_{\theta})+  \hat{\mathbf{R}}_{\mathcal{P}^{\text{mix}}_{\delta}}(f_{\theta},g_{\phi}) + \hat{\mathbf{R}}_{\mathcal{P}^{\text{mix}}_{\delta}}(f_{\theta})- \hat{\mathbf{R}}_{\mathcal{P}^{\text{mix}}_{\delta}}(f_{\theta})|
\end{equation}
using the triangle inequality on the absolute value we have:
\small
\begin{equation}
    \mathcal{L}(\theta)\leq 2 \mathbf{R}_{\mathcal{P}}(f_{\theta}) + | \hat{\mathbf{R}}_{\mathcal{P}_{\delta}}(f_{\theta}) -\mathbf{R}_{\mathcal{P}}(f_{\theta})| + |   \hat{\mathbf{R}}_{\mathcal{P}^{\text{mix}}_{\delta}}(f_{\theta}) -\mathbf{R}_{\mathcal{P}}(f_{\theta})| +| \hat{\mathbf{R}}_{\mathcal{P}^{\text{mix}}_{\delta}}(f_{\theta},g_{\phi})-  \hat{\mathbf{R}}_{\mathcal{P}^{\text{mix}}_{\delta}}(f_{\theta})|
\end{equation}
\normalsize

Let us first focus on the last term of the sum and use the integral absolute value inequality:
\begin{equation}
| \hat{\mathbf{R}}_{\mathcal{P}^{\text{mix}}_{\delta}}(f_{\theta},g_{\phi})-  \hat{\mathbf{R}}_{\mathcal{P}^{\text{mix}}_{\delta}}(f_{\theta})|
\leq \int | l(f_{\theta}(x),g_{\phi}(x)) - l(f_{\theta}(x),y) |d\mathcal{P}^{\text{mix}}_{\delta}(x,y)
\end{equation}
Then thanks to the triangle inequality on $l$ we have
\begin{equation}
| \hat{\mathbf{R}}_{\mathcal{P}^{\text{mix}}_{\delta}}(f_{\theta},g_{\phi})-  \hat{\mathbf{R}}_{\mathcal{P}^{\text{mix}}_{\delta}}(f_{\theta})|
\leq \int  l(y,g_{\phi}(x)) d\mathcal{P}^{\text{mix}}_{\delta}(x,y) = \hat{\mathbf{R}}_{\mathcal{P}^{\text{mix}}_{\delta}}(g_{\phi})
\end{equation}
Now let us focus on the second term:
It can be rewritten:
\begin{equation}
| \hat{\mathbf{R}}_{\mathcal{P}_{\delta}}(f_{\theta}) -\mathbf{R}_{\mathcal{P}}(f_{\theta})|
= | \int  l(f_{\theta}(x),y)( \mathcal{P}_{\delta}(x,y) -\mathcal{P}(x,y))   dxdy| 
\end{equation}
using the integral absolute value inequality:
\begin{equation}
| \hat{\mathbf{R}}_{\mathcal{P}_{\delta}}(f_{\theta}) -\mathbf{R}_{\mathcal{P}}(f_{\theta})|
\leq \int  l(f_{\theta}(x),y)| \mathcal{P}_{\delta}(x,y) -\mathcal{P}(x,y)|   dxdy 
\end{equation}
Then we have:
\begin{equation}
| \hat{\mathbf{R}}_{\mathcal{P}_{\delta}}(f_{\theta}) -\mathbf{R}_{\mathcal{P}}(f_{\theta})|
\leq M \int | \mathcal{P}_{\delta}(x,y) -\mathcal{P}(x,y)|   dxdy 
\end{equation}
with $M=\sup(l(f_{\theta}(x),y))$, hence we have:
\begin{equation}
| \hat{\mathbf{R}}_{\mathcal{P}_{\delta}}(f_{\theta}) -\mathbf{R}_{\mathcal{P}}(f_{\theta})|
\leq M \| \mathcal{P}_{\delta} -\mathcal{P}\|_1
\end{equation}
similarly we have:

\begin{equation}
|   \hat{\mathbf{R}}_{\mathcal{P}^{\text{mix}}_{\delta}}(f_{\theta}) -\mathbf{R}_{\mathcal{P}}(f_{\theta})|
\leq M \| \mathcal{P}^{\text{mix}}_{\delta} -\mathcal{P}\|_1
\end{equation}

\end{proof}

This implies that the quality of the DCNN is bounded by the accuracy of the teacher. It is also bounded by how much the mixing strategy can sample the true distribution of the dataset. Finally, the distribution of the training data with respect to the true data distribution also plays an important role.

\section{Extra Experiments}

This section adds some complementary results on the Cityscape-C experiments. Moreover, to have a better understanding of Superpixel-mix, we conduct an ablation study on its parameters. We also complete the SSL experiments by adding results to the Pascal dataset. %

\subsection{Complement Cityscapes-C}

In semantic segmentation, the 
DCNN must be reliable to distributional shift uncertainty. %
To check that, we generate Cityscapes-C dataset based on the code of Hendrycks et al. \cite{hendrycks2019benchmarking}. Note that Cityscapes-C is composed of 16 types of pertubutions. Here is the list of all perturbations: Gaussian noise, shot noise, impulse noise, defocus blur, frosted, glass blur, motion blur, zoom blur, snow, frost, fog, brightness, contrast, elastic, pixelate, and JPEG.  In addition, each type comprises five levels of severity. Playing with these five levels is essential since we can check how an algorithm evolves with the severity.

\begin{figure*}[h]
\centering

\includegraphics[width=0.8\linewidth]{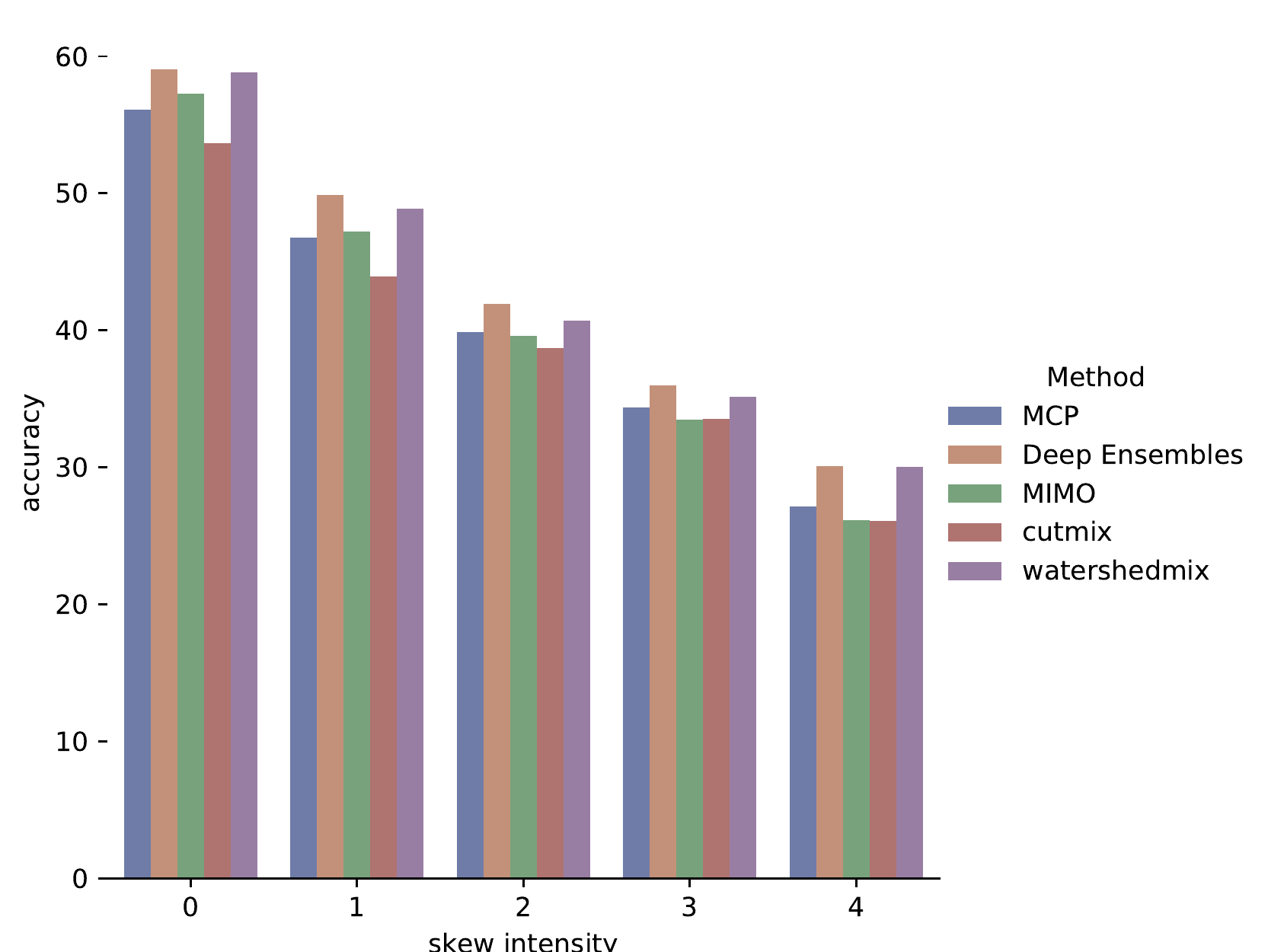}
\caption{
Results on Cityscapes-C dataset's mIoU for the different level of noise intensity.
}
\label{fig:Cityscapes-C}
\end{figure*}

 In Figure \ref{fig:Cityscapes-C} we illustrate the mIoU of different approaches for the different levels of noise. We can see that Superpixel-mix tends to be resistant to high level of noise, while except for Deep Ensembles, competitors have difficulties. This property is interesting since it shows that Superpixel-mix is {more} reliable 
even in highly 
 uncertain environments.

\subsection{Ablation studies on Superpixels}\label{ablation_superpixel}

Our algorithm has two parameters {to set}:  the number of superpixels, and the proportion of selected superpixels {used as masks for mixing.}
{In this section we study the impact of the choice of these parameters over the peformance downstream. To this end, we conduct an ablation on Cityscapes over the same split of 744 images.}

Our first study is related to the number of superpixels in Superpixel-mix {and report results in Table~\ref{table:Abblation2_nb_sp}}.  
We can see that the performance increases with the number of superpixels up to 200. 
{After this point, the mIoU score decreases.}
The number of superpixels is directly linked to their size. It is also connected the number of salient edges that will be kept from the original images. Hence, we can deduce that most of the true edges are discarded in the case of a small number of superpixels, leading to low performances. While in the case where we have a high number of superpixels, we might have an over-segmentation that 
{likely leads to a high granularity non-informative masks that prevent learning representations for objects and object parts.}
{In such cases, performance is lower.}

Our second study is linked to the proportion of selected superpixels. The results of this survey are in Table \ref{table:Abblation1_prop}. {We can see that the performance is stable across the range of different values.}

\begin{table}[!t]
\begin{center}
 \scalebox{0.90}
 {
\begin{tabular}{l| c  c  c  c  c  c  c  c  }
\toprule
 Nb. superpixels &  20 &  50 & 100  &  200&  500& 1000 \\

 mIoU & 63.81  \%& 64.47\% & 65.16\%  &65.0 \%& 64.16  \%& 64.2\%  \\
\bottomrule
\end{tabular}
}
\end{center}
\caption{Ablation study results on the number of superpixels on Cityscapes dataset. All DCNNs are trained on the same split of 1/7 image set under the same conditions.}
\label{table:Abblation2_nb_sp}
\end{table}

\begin{table}[!t]
\begin{center}
 \scalebox{0.8}
 {
\begin{tabular}{l | c c  c  c c c  c  c c}
\toprule
 Proportion &  0.1 &  0.2 & 0.3 &0.4 & 0.5 & 0.6 &  0.7 & 0.8  & 0.9\\
 Performance (mIoU) & 65.59  \%& 65.54\% & 65.43\%  & 65.0\% & 65.29\% & 65.63  \%& 64.9\% & 65.59 & 65.41\\
\bottomrule
\end{tabular}
}
\end{center}
\caption{Ablation study results on proportion of chosen superpixels on Cityscapes dataset. All DCNNs are trained on the same split of 1/7 image set under the same conditions.}
\label{table:Abblation1_prop}
\end{table}

\subsection{Semi-supervised experiments on Pascal}

{We evaluate our method for semi-supervised semantic segmentation on the Pascal VOC 2012 dataset~\cite{pascal-voc-2012}. We compare against top existing methods, following the common protocol from prior works~\cite{adversarial_bmvc_2018, cutmix_SSL_seg_BMVC_2020}, i.e., four sets of labeled data: 1/100 (106 images),	1/50 (212 images),	1/20 (529 images), 1/8 (1323 images). We report results in Table~\ref{table:Pascal}. All methods use the same training data split\footnote{\url{https://github.com/Britefury/cutmix-semisup-seg/tree/master/data/splits/pascal_aug}}, however, in contrast with prior works that report results only from a single training run, we conduct six different runs to assess the stability of our approach and report mean mIoU scores and standard deviation. }

\begin{table*}[ht!]
\begin{center}
\scalebox{0.6}
 {
\begin{tabular}{l l l l l}
\toprule
\textbf{Labeled samples} & \textbf{1/100 (106)} & \textbf{1/50 (212)} & \textbf{1/20 (529)} & \textbf{1/8 (1323)} \\
\midrule
Adversarial~\cite{adversarial_bmvc_2018} & - &  57.2\% & 64.7\% & 69.5\% \\
s4GAN~\cite{s4GAN_pami_2019} & - & 63.3\% & 67.2\% & 71.4\% \\
Cutout~\cite{cutout_arXiv_2017} & 48.73\% & 58.26\% & 64.37\% &  66.79\% \\
Cutmix~\cite{cutmix_SSL_seg_BMVC_2020}\footnote{note that we trained cutmix with COCO pretraining} & 57,01\% & 65,99\% & 68,3\% &  71,2\% \\
Classmix~\cite{classmix_arXiv_2020} & 54.18\% & 66.15\% & 67.77\% &  71.00\% \\
DMT\cite{feng2020semi} & 61.6\% &  65.5\% & 69.3\% & 70.7\% \\
\midrule
Baseline(*) & 42.47\% & 55.69\% & 61.36\% & 67.14\% \\
Superpixel-mix (ours) & \textbf{57,69\% $\pm$ 0,53} ($\uparrow$ 15.22\%) &\textbf{ 66,73\% $\pm$ 0,54 ($\uparrow$ 11.04\%)} & \textbf{69.87\% $\pm$ 0,39}
($\uparrow$ 8.51\%) & \textbf{72,04\% $\pm$ 0,40 ($\uparrow$ 4.9\%)}\\
\bottomrule
\end{tabular}
}
\end{center}
\caption{Performance (mIoU) on Pascal VOC 2012~\cite{pascal-voc-2012} on the validation set, which is computed over official split used by  \cite{cutmix_SSL_seg_BMVC_2020}. For Superpixel-mix we report scores averaged over six different runs.}
\label{table:Pascal}
\end{table*}

\subsection{Semi-supervised experiments on ISIC 2017}

We evaluate our method for semi-supervised semantic segmentation on the ISIC skin lesion segmentation dataset \cite{codella2018skin}. We compare ours against the top existing methods, following the common protocol from prior works~\cite{adversarial_bmvc_2018, cutmix_SSL_seg_BMVC_2020}. We use 50 out of the 2000 training images and scaled them to $ 248 \times 248$. Then we apply a random crop of $ 224 \times 224$ with random flips and rotations, and uniform scaling in the range from 0.9 to 1.1.
We report results in Table~\ref{table:ISIC}. The results are averaged on 5 different splits. 
For this dataset, similarly to ~\cite{adversarial_bmvc_2018, cutmix_SSL_seg_BMVC_2020,li2018semi}, we use DenseUNet-161 pretrained on Imagenet.

\begin{table*}[ht!]
\begin{center}
\scalebox{0.6}
 {
\begin{tabular}{l l }
\toprule
\textbf{Labeled samples} & \textbf{(50)} \\
\midrule
Self ensemble~\cite{li2018semi} & \textbf{75.31\%} \\
Cutout~\cite{cutout_arXiv_2017} & 68.76\%  \\

Cutmix~\cite{cutmix_SSL_seg_BMVC_2020} & 74.57\%  \\
\midrule
Baseline(*) & 67.64\%\\
Superpixel-mix (ours) & \textbf{74.53\% $\pm$ 1,23} ($\uparrow$ 6.89\%) \\
\bottomrule
\end{tabular}
}
\end{center}
\caption{Performance (mIoU) on  ISIC skin lesion segmentation dataset \cite{codella2018skin} on the validation set. The results are averaged over 5 splits.}
\label{table:ISIC}
\end{table*}

\section{Novel dataset Out of Context Cityscapes (OC-Cityscapes)}
\begin{figure*}[h]
\centering
\begin{tabular}{cc}
\includegraphics[width=0.4\linewidth]{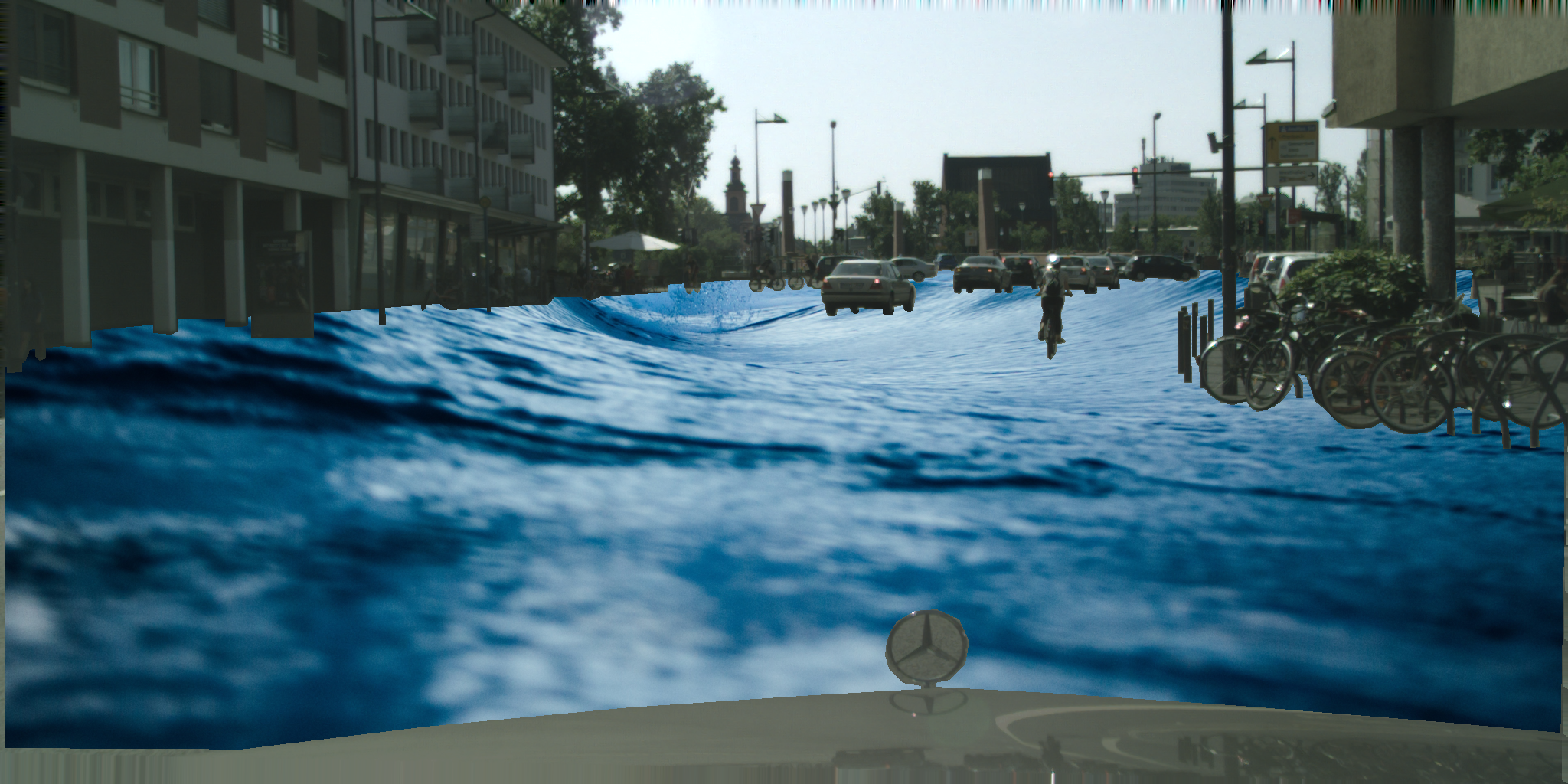}&
\includegraphics[width=0.4\linewidth]{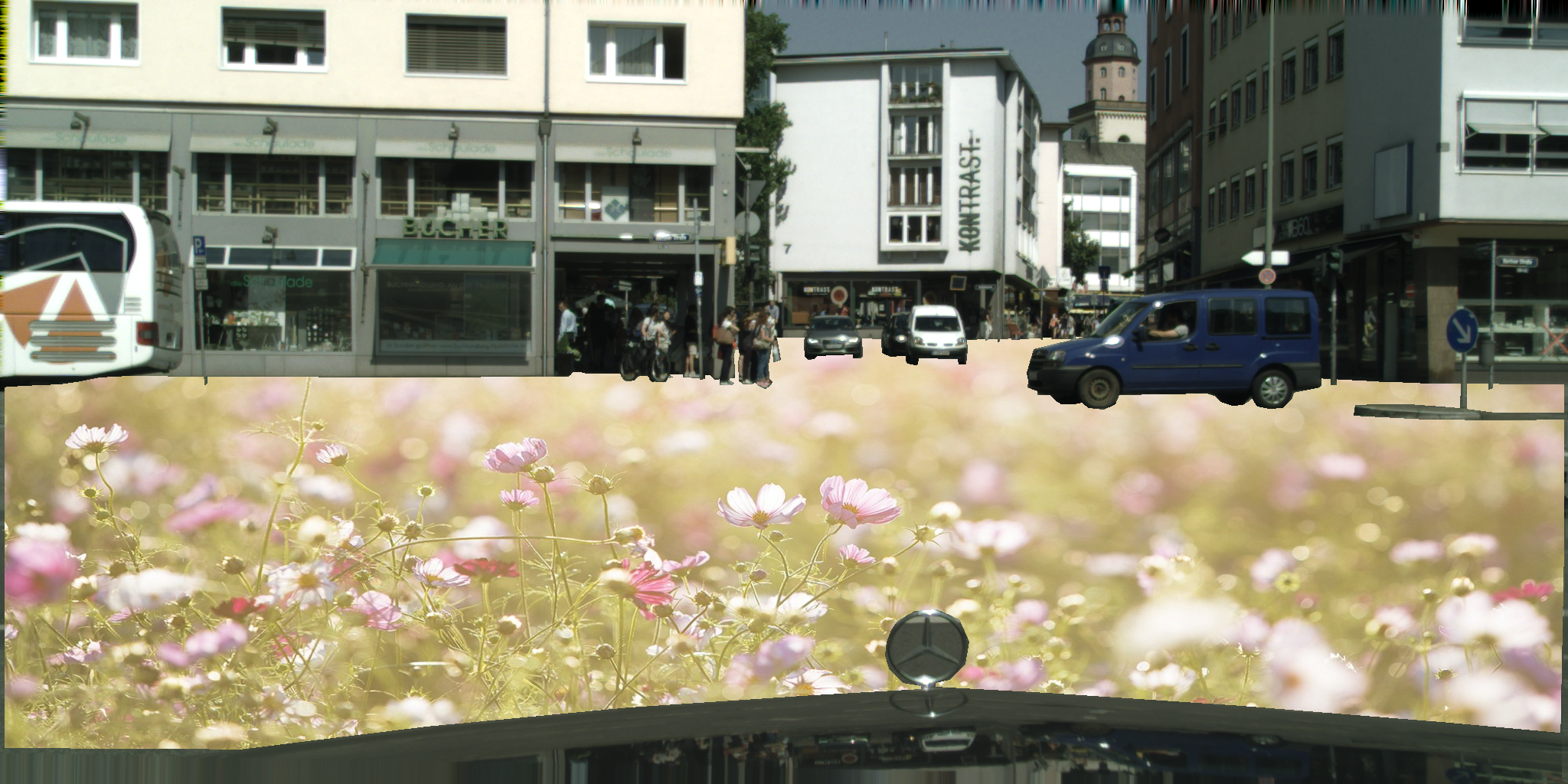}\\
\includegraphics[width=0.4\linewidth]{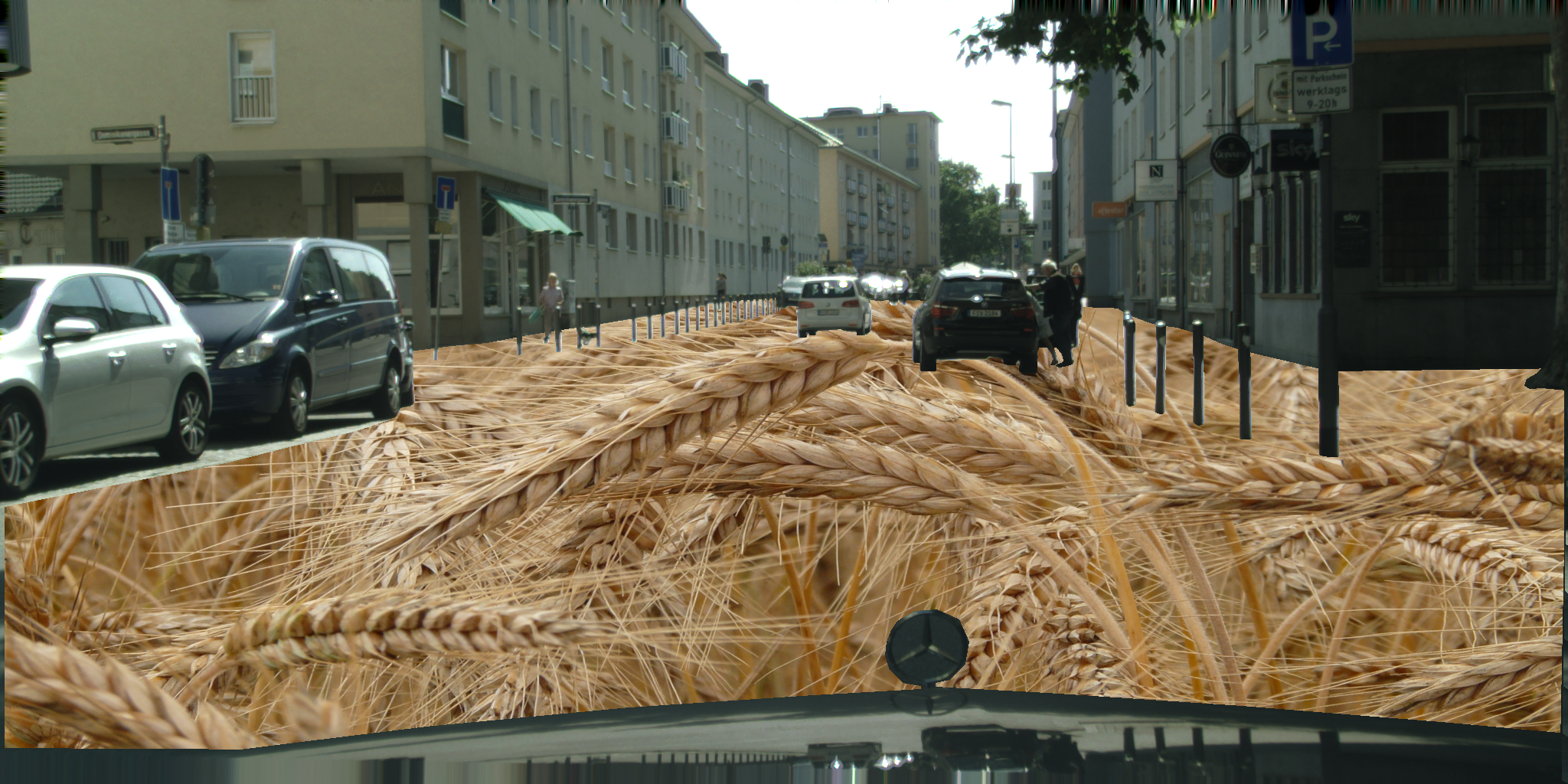}&
\includegraphics[width=0.4\linewidth]{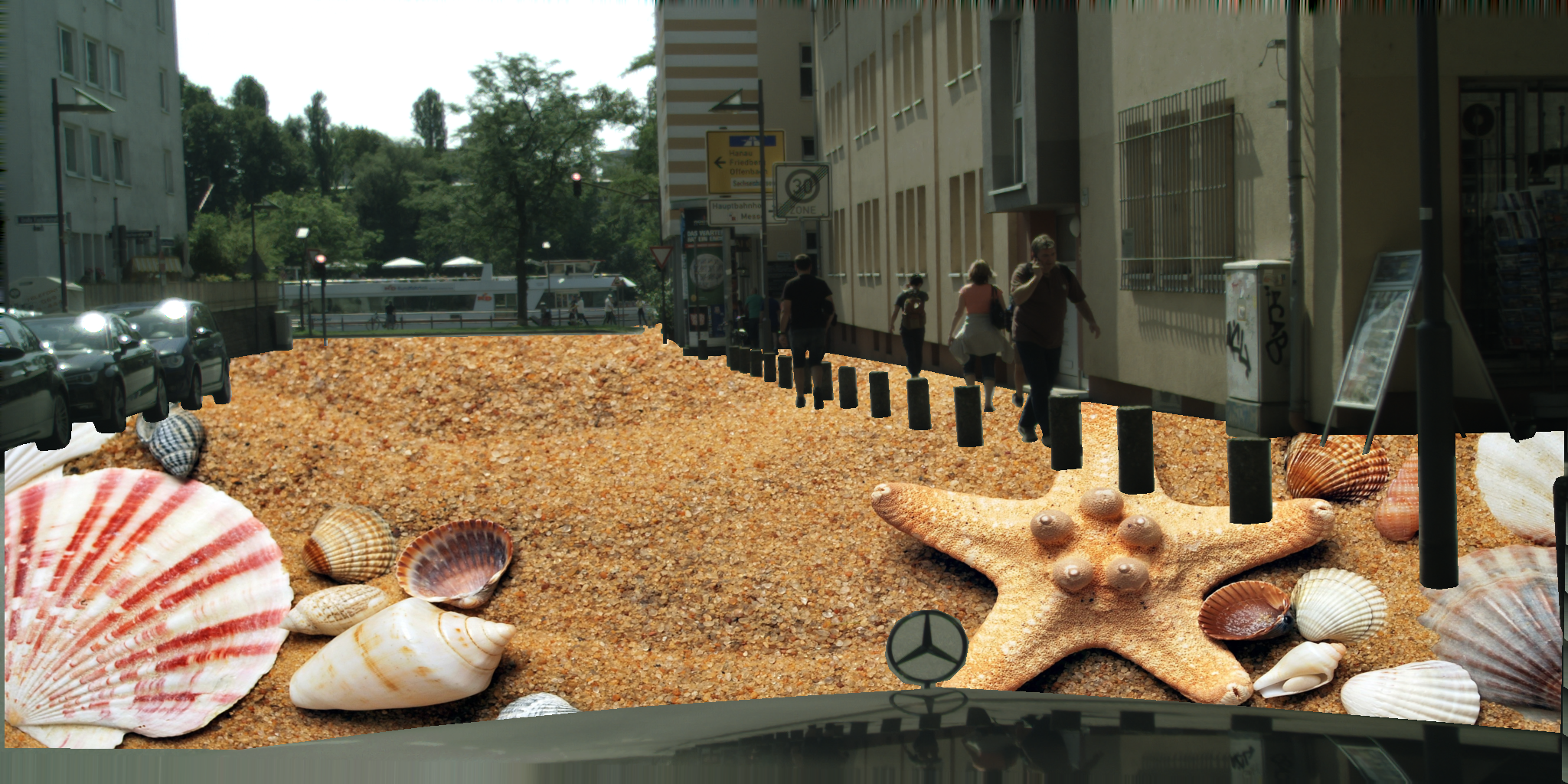}\\
\includegraphics[width=0.4\linewidth]{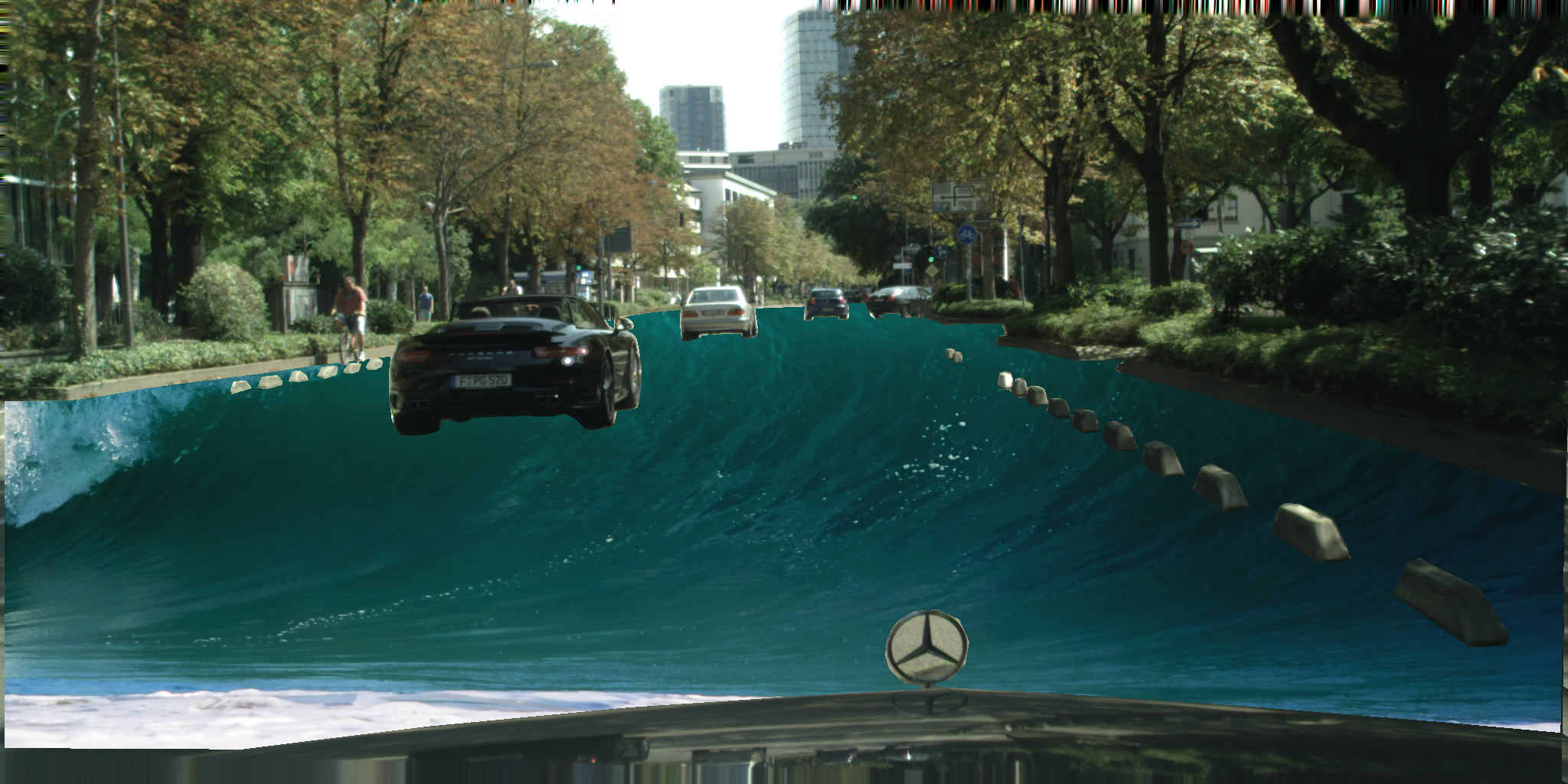}&
\includegraphics[width=0.4\linewidth]{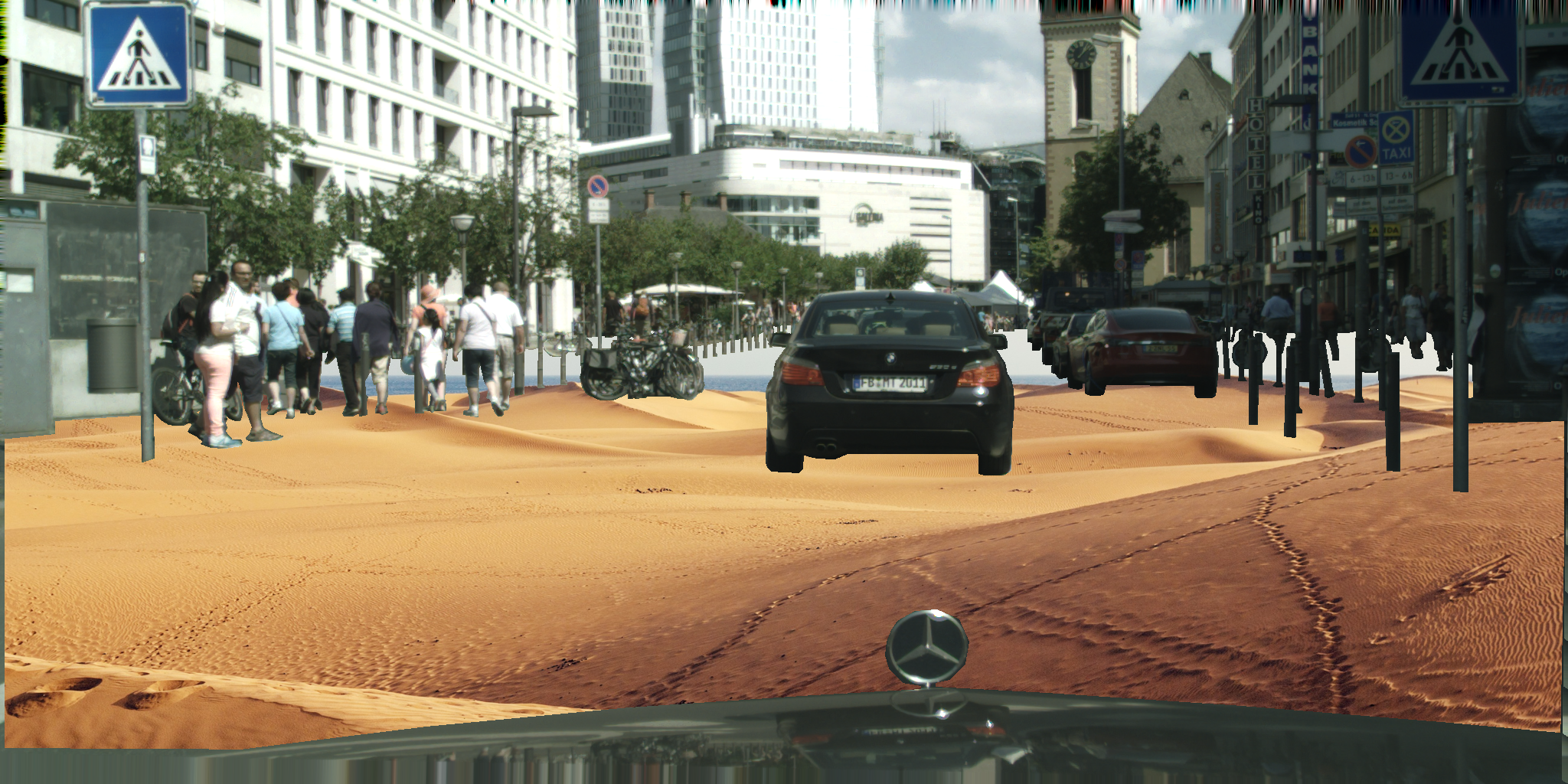}\\
\includegraphics[width=0.4\linewidth]{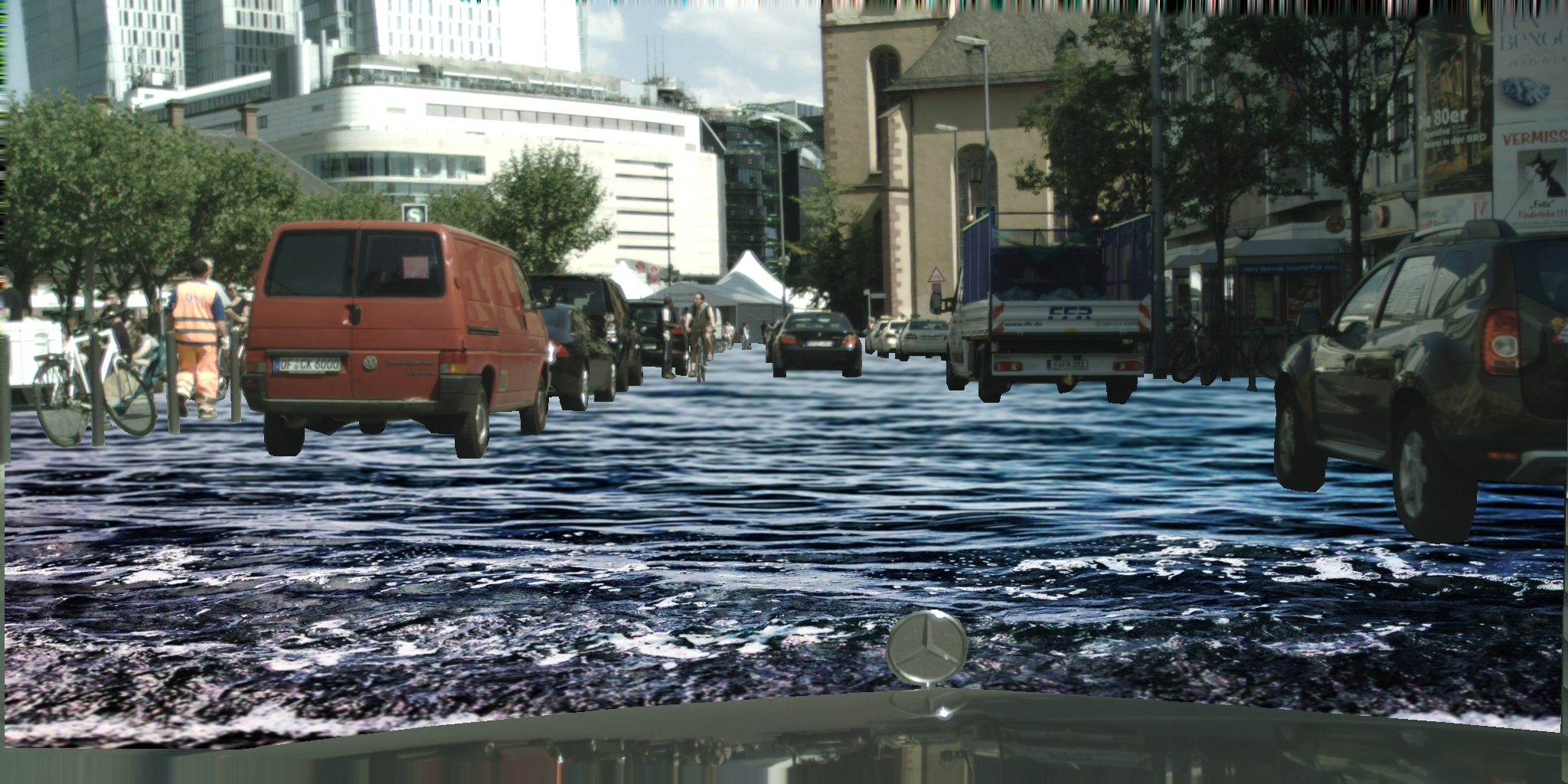}&
\includegraphics[width=0.4\linewidth]{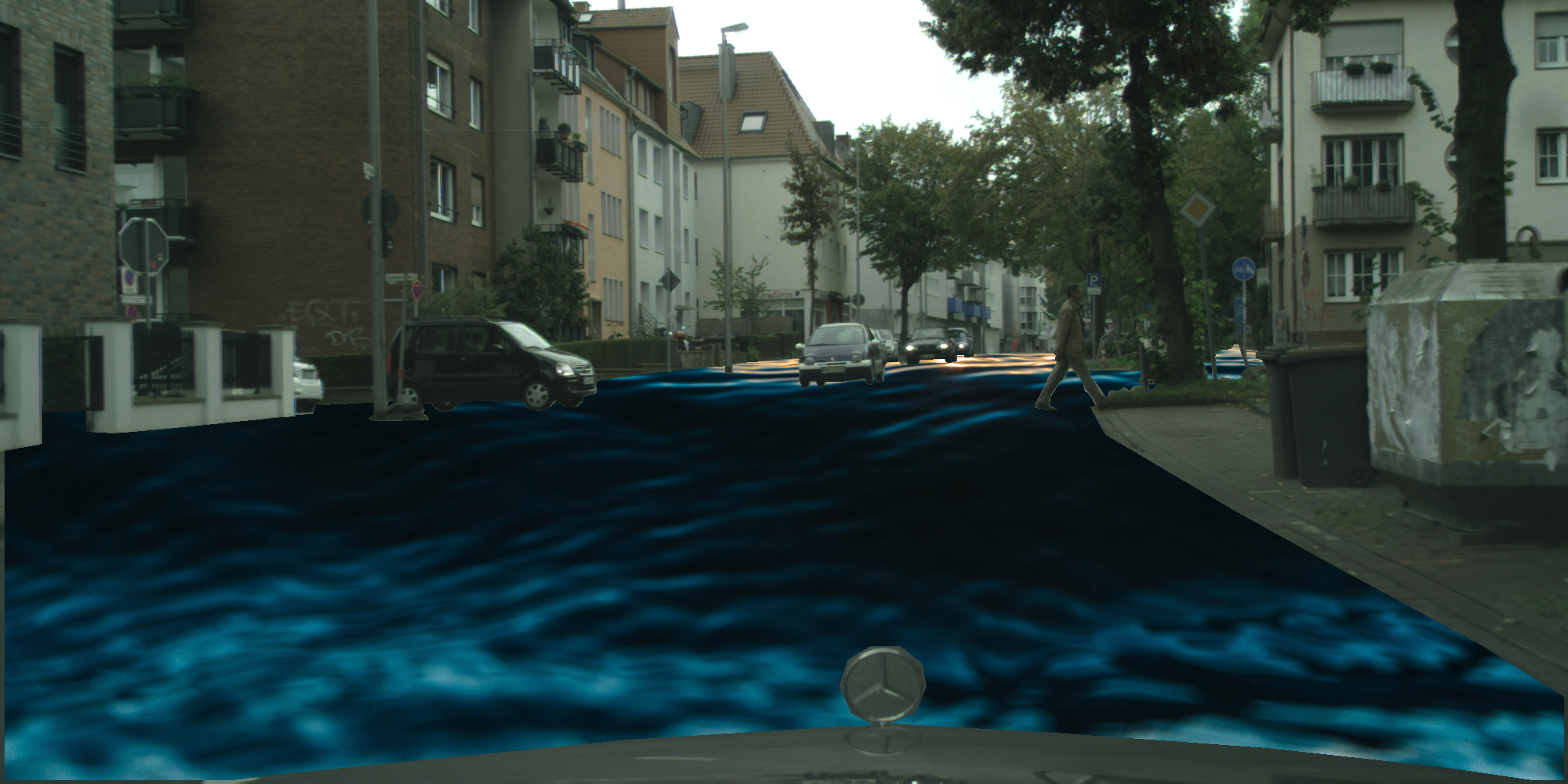}\\
\includegraphics[width=0.4\linewidth]{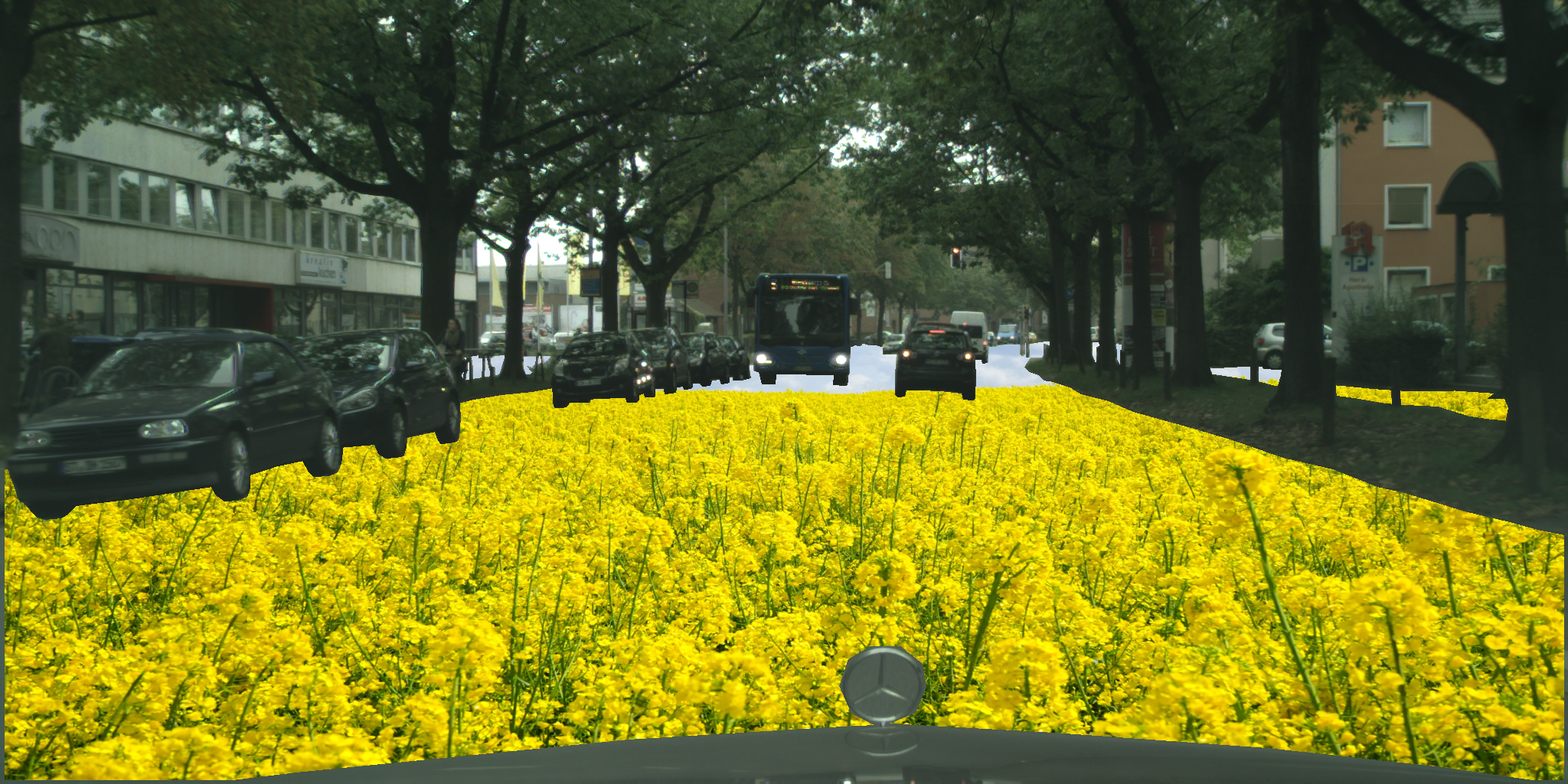}&
\includegraphics[width=0.4\linewidth]{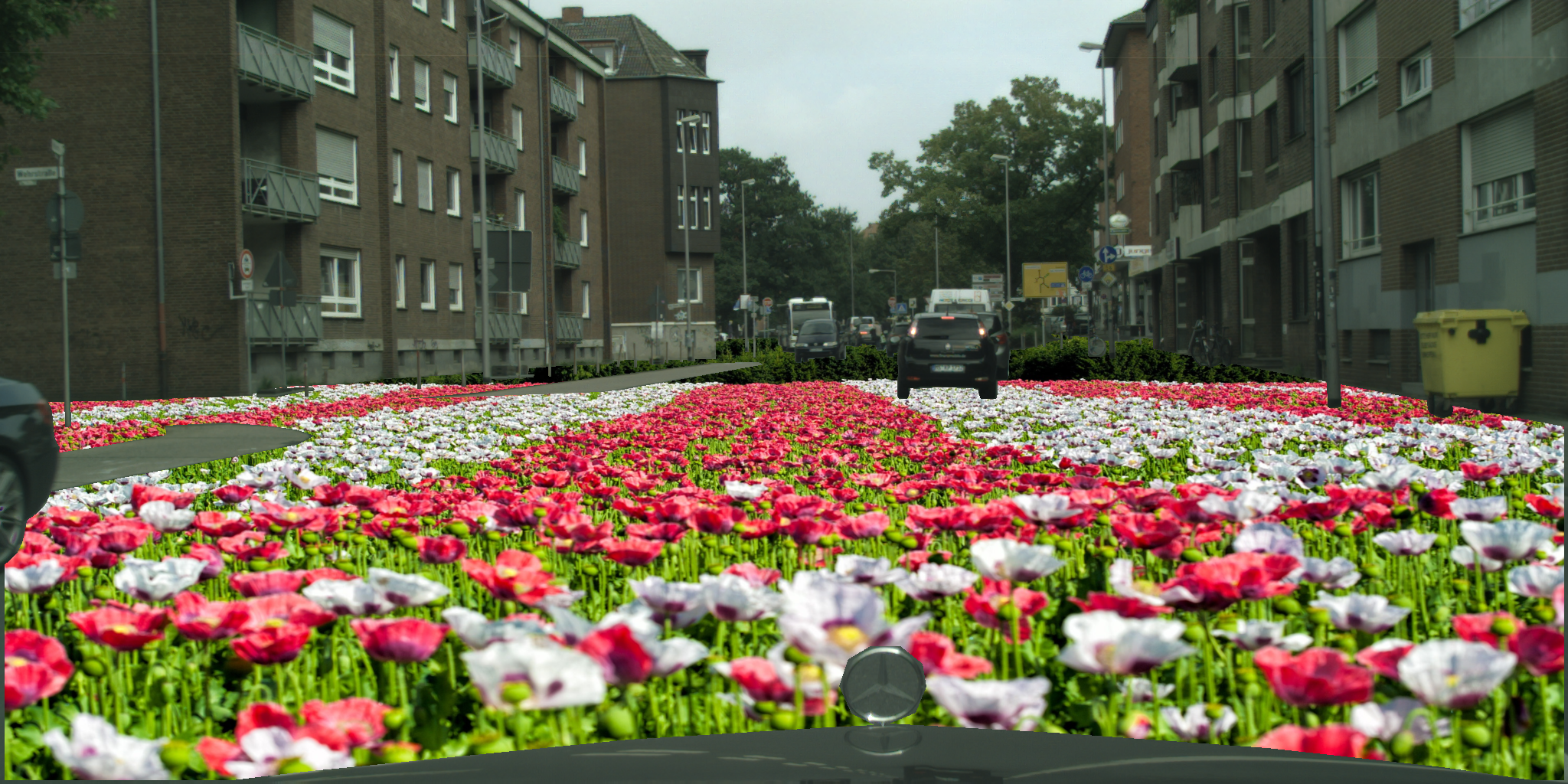}
\end{tabular}
\caption{Illustration of some images of OC-Cityscapes dataset}
\label{fig:dataset}
\end{figure*}

In Figure \ref{fig:dataset}, we illustrate 
a few example images from the contextual free Cityscape dataset used in the unbiaising DCNN experiment. To build this dataset, we replace the pavements and roads with natural landscapes such as sea, forest, desert background. 
\ab{These extreme settings allow to better identify and assess potential contextual biases of semantic segmentation models.}
The dataset will be made publicly available after the anonymity period.

\newpage

\section{Novel dataset Out of Context Cityscapes (OC-Cityscapes)}
In Figure \ref{fig:dataset}, we illustrate 
a few example images from the contextual free Cityscape dataset used in the unbiaising DCNN experiment. To build this dataset, we replace the pavements and roads with natural landscapes such as sea, forest, desert background. 
\ab{These extreme settings allow to better identify and assess potential contextual biases of semantic segmentation models.}

\section{Implementation details}

In this section, we provide the hyper-parameters that are used in the semantic-segmentation experiments. Our code is implemented 
\ab{in PyTorch~\cite{paszke2019pytorch}.}
The code will be made publicly available after the anonymity period.

\begin{table}[h!]
\begin{center}
\scalebox{0.85}
{
\begin{tabular}{l|c|c|c}
\toprule
  \ab{\textbf{Hyper-parameter}} &   \textbf{StreetHazards} & \textbf{Cityscape}    & \textbf{ISIC 2017}     \\ 
\midrule
Architecture         &Deeplab v3+ & Deeplab v3+ & DenseUNet-161   \\ 
\midrule
output stride       &16 &8 &-\\ 
\midrule
learning rate         &0.1 &0.1 &0.1 \\ 
 \midrule
batch size        &4 & 16 & 8  \\ 
 \midrule
number of train epochs  & 25 & 25 & 25  \\ 
\midrule
 weight decay  &0.0001 & 0.0001& 0.0001 \\ 
 \midrule
 SyncEnsemble BN        & False & False  & False \\ 
  \midrule
 random crop of training images    & None & 768 & 224 \\ 
\bottomrule
\end{tabular}
}
\end{center}
\caption{
\textbf{Hyper-parameter configuration used in the fully supervised semantic segmentation experiments .
}
}\label{table:tab3}
\end{table}

\begin{table}[h!]
\begin{center}
\scalebox{0.85}
{
\begin{tabular}{l|c|c}
\toprule
  \ab{\textbf{Hyper-parameter}} &   \textbf{Cityscape} & \textbf{Pascal}        \\ 
\midrule
Architecture         &Deeplab v2 & Deeplab v2  \\ 
\midrule
output stride       &16 &16 \\ 
\midrule
learning rate         &2.5e-4 &2.5e-4\\ 
 \midrule
batch size        &2 & 5 \\ 
 \midrule
number of training iteration  & 40000 & 40000\\ 
\midrule
 weight decay  &5e-4 & 5e-4  \\ 
 \midrule
 SyncEnsemble BN        & True & True \\ 
\bottomrule
\end{tabular}
}
\end{center}
\caption{
\textbf{Hyper-parameter configuration used in the semi supervised semantic segmentation experiments .
}
}\label{table:tab31}
\end{table}

\end{document}